\documentclass[10pt,twocolumn,letterpaper]{article}

\usepackage{cvpr}              
\usepackage{multirow}
\definecolor{cvprblue}{rgb}{0.21,0.49,0.74}
\usepackage[pagebackref,breaklinks,colorlinks,allcolors=cvprblue]{hyperref}

\def\paperID{7120} 
\def\confName{CVPR}
\def\confYear{2026}

\title{No Labels, No Look-Ahead: Unsupervised Online Video Stabilization with Classical Priors}

\author{
Tao Liu$^{1}$\thanks{Equal contribution} \quad
Gang Wan$^{1}$ \quad
Kan Ren$^{1}$\thanks{Corresponding author} \quad
Shibo Wen$^{2}$\footnotemark[1]\\
$^1$School of Electronic and Optical Engineering, Nanjing University of Science and Technology\\
$^2$College of Geoexploration Science and Technology, Jilin University\\
{\tt\small liutao23@njust.edu.cn, wangang@njust.edu.cn, K.ren.@njust.edu.cn, wensb23@mails.jlu.edu.cn}
}

\begin{document}
\maketitle
\begin{abstract}
We propose a new unsupervised framework\footnotemark for online video stabilization. Unlike methods based on deep learning that require paired stable and unstable datasets, our approach instantiates the classical stabilization pipeline with three stages and incorporates a multithreaded buffering mechanism. This design addresses three longstanding challenges in end-to-end learning: limited data, poor controllability, and inefficiency on hardware with constrained resources. Existing benchmarks focus mainly on handheld videos with a forward view in visible light, which restricts the applicability of stabilization to domains such as UAV nighttime remote sensing. To fill this gap, we introduce a new multimodal UAV aerial video dataset (UAV-Test). Experiments show that our method consistently outperforms state-of-the-art online stabilizers in both quantitative metrics and visual quality, while achieving performance comparable to offline methods.
\end{abstract}
\footnotetext{\url{https://github.com/liutao23/LightStab.git}.}

\section{Introduction}
\label{sec:intro}
Video stabilization is a common requirement in both amateur and professional video capture, aiming to suppress unintended camera shake and enhance visual quality. To avoid the high cost and operational complexity of physical stabilizers, researchers have proposed a wide range of algorithmic solutions. Classical approaches typically follow a three-stage pipeline: motion estimation, motion smoothing, and stabilized frame generation. Based on the dimensionality of the motion model, these methods can be classified as 2D, 2.5D, or 3D. 2D methods adopt affine transformations~\cite{r1}, homographies~\cite{r2}, or optical flow~\cite{r3,r4}, achieving high efficiency but struggling with parallax and depth variations. 2.5D approaches~\cite{r5,r6} incorporate limited 3D cues from feature trajectories to mitigate parallax, but remain computationally expensive. 3D methods leverage depth sensors~\cite{r7}, reconstructed point clouds~\cite{r8}, or epipolar geometry~\cite{r9} to address complex scenes effectively, but entail significant computational overhead.

Deep learning enables end-to-end stabilization by directly generating stabilized frames from shaky inputs. However, such frameworks are often difficult to interpret or control and typically rely on large paired datasets of stable/unstable videos. In practice, unavoidable temporal and spatial misalignments make available data prone to parallax effects~\cite{r10}, limited scenario coverage~\cite{r11}, and synthetic-to-real gaps~\cite{r12}. Recent studies, therefore, explore a decoupled design. DUT~\cite{r13} augments the classical pipeline with neural motion propagation and smoothing, achieving strong stabilization. Nevertheless, it remains constrained by its network architecture~\cite{r14} and a global smoothing strategy, and therefore operates offline. Another line of work~\cite{r15} leverages high-quality motion-estimation networks~\cite{r16} and focuses on optimizing the smoothing module, thereby enabling online stabilization. Yet, the lack of open-source motion estimators and reduced robustness in complex scenes still pose significant challenges.

\begin{figure*}[t]
\centering
\includegraphics[width=0.99\linewidth]{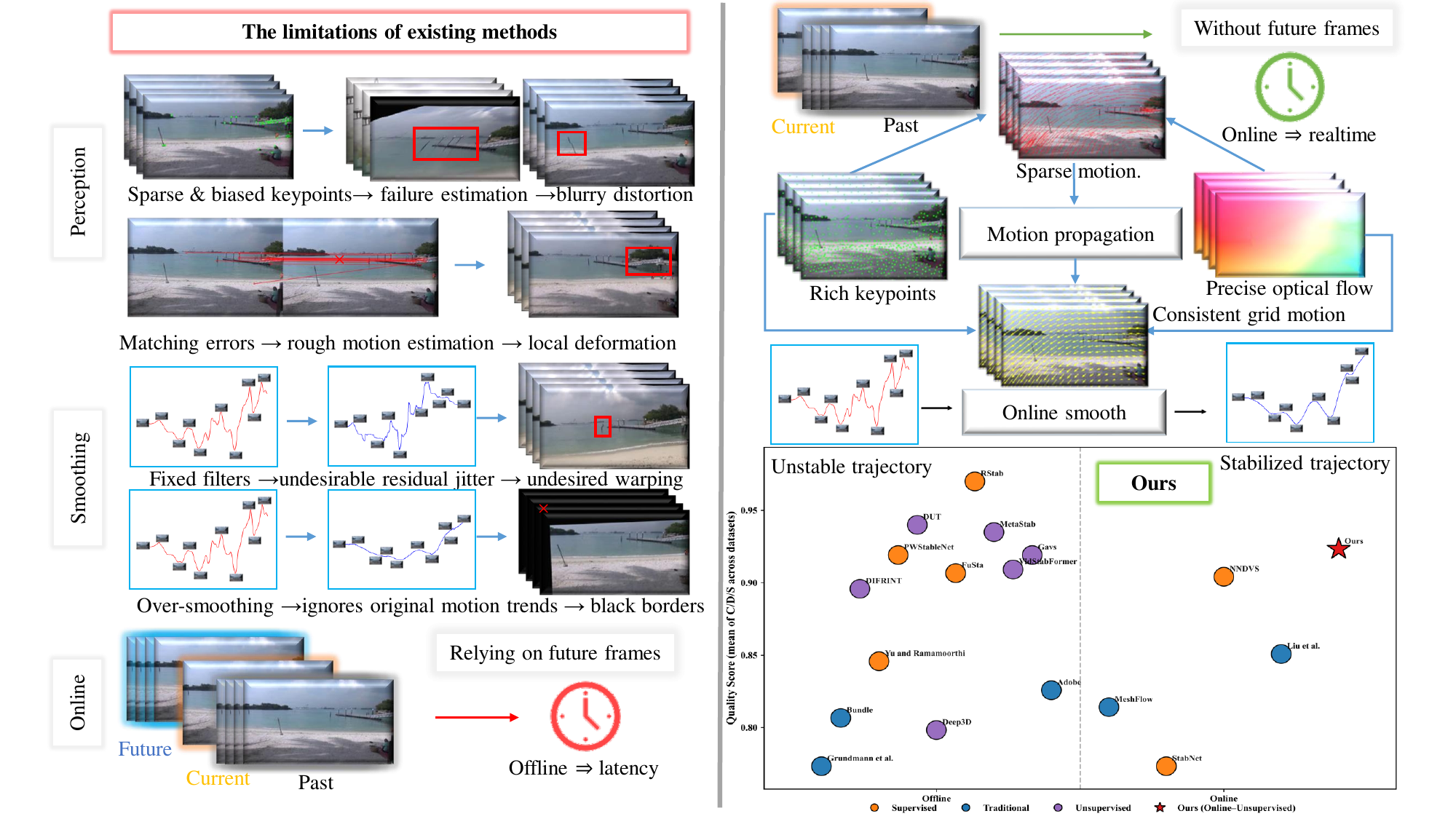}
\caption{Limitations of existing methods and an overview of our approach.
(a) Traditional pipelines rely on sparse, spatially biased keypoints, leading to inaccurate motion estimation and blur/distortion in stabilized frames.
(b) Over-smoothing and fixed filtering ignore scene motion trends, resulting in residual jitter, black borders, and warping in the output video.
(c) Our approach uses dense, evenly distributed keypoints together with accurate optical flow to propagate motion estimates across frames; it operates online without future frames, enabling real-time processing.
(d) In the bottom-right chart, our method (red star) outperforms both traditional offline methods and recent online methods in stabilization quality, achieving state-of-the-art performance for online unsupervised stabilization.}
\label{fig:motivation}
\end{figure*}

Our literature review and empirical analysis highlight three key limitations common to both classical and learning-based stabilizers:

\begin{itemize}
\item \textbf{Perception Limitations:} Classical methods rely on handcrafted keypoint detectors~\cite{r17,r18,r19}, which struggle with weak textures, occlusions, and large motions, leading to biased motion estimation~\cite{r20}. Although deep learning methods~\cite{r21} improve robustness, they lack interpretability and controllability, limiting their adaptability to real-world scenarios.
\item \textbf{Smoothing Limitations:} Classical methods employ fixed smoothing strategies~\cite{r22}, which fail to generalize well, resulting in residual jitter~\cite{r23}. Learning-based smoothing methods lack geometric interpretability~\cite{r24}, thereby risking over-smoothing or distortion~\cite{r10}.
\item \textbf{Online Processing Limitations:} Many stabilizers, both classical and learning-based, are not designed for online processing and rely on offline aggregation or batch processing, introducing delays. Additionally, learning-based methods require large labeled datasets and significant computational resources~\cite{r10}, hindering online deployment.
\end{itemize}

As shown in Fig.~\ref{fig:motivation}, existing stabilizers face challenges in motion perception and trajectory smoothing, which often lead to artifacts under complex conditions, and are mostly limited to offline processing. To address these challenges, we propose the following solutions. For \textbf{Motion Perception}, we combine multi-detector collaboration with keypoint uniformization to enhance robustness, reduce optical flow overhead, and introduce an efficient motion propagation network to maintain global motion consistency. In terms of \textbf{Trajectory Smoothing}, we present an online dynamic-kernel optimization method that avoids fixed filters and future-frame dependencies, enabling online stabilization with improved control. For \textbf{System Design}, we implement a multi-threaded asynchronous pipeline that decouples and parallelizes core modules, reducing latency and enabling high-frame-rate processing.

In summary, our framework combines robust motion perception, interpretable trajectory smoothing, and efficient system design into a \textbf{simulation-informed stabilization chain}, which overcomes existing limitations in perception, analysis, and online performance. The main contributions of this work are as follows:

\begin{itemize}
\item We propose a novel unsupervised online video stabilization model that removes the need for paired training data while enabling online processing of unstable videos.
\item We introduce \textbf{UAV-Test}, a realistic multimodal dataset of unstable aerial videos covering diverse drone-shake scenarios, intended as a benchmark for related research, to be released upon acceptance.
\item Extensive experiments on public and newly collected datasets show that our method surpasses state-of-the-art online stabilizers in quality and achieves performance on par with offline approaches.
\end{itemize}

\section{Related Work}
\label{chap:RelatedWork}

\subsection{Video Stabilization}
Traditional video stabilization methods \cite{r1,r20} suppress screen jitter by estimating camera motion trajectories and applying smoothing techniques, or by using adaptive sparse grids for real-time performance \cite{r23}. However, these methods often rely on manually designed feature detectors, which may lead to uneven keypoint distribution, require offline processing in most cases, and exhibit limited robustness in complex scenes. With the development of deep learning, the end-to-end, data-driven paradigm has gradually become dominant. StabNet \cite{r10} was the first to propose a supervised online stabilization framework based on convolutional neural networks (CNNs), followed by PWStableNet \cite{r21} and Yu et al. \cite{r4}, which introduced pixel-level 2D deformation fields for more precise motion correction. FuSta \cite{r27} improved stabilization performance by integrating multi-scale spatial and feature information, while DIFRINT \cite{r24} effectively reduced cropping artifacts through unsupervised inter-frame interpolation. Subsequently, DUT \cite{r13} achieved stabilization without manual annotation by utilizing motion priors or trajectory consistency. NNDVS \cite{r15} achieved high-quality, low-latency video stabilization using existing motion estimation frameworks. FastStab \cite{r70} enabled fast offline 2D stabilization through optical flow guidance. To further enhance temporal consistency and scene adaptability, VidStabFormer \cite{r64} and MetaStab \cite{r67} introduced self-supervised and meta-learning strategies, respectively. In recent years, 3D-based stabilization techniques have emerged: Deep3D \cite{r7} jointly estimates depth and pose; DeepFused \cite{r68} integrates optical flow and gyroscope data to predict virtual 3D camera trajectories; RStab \cite{r28} employs a neural rendering technique with adaptive modules; and Gavs \cite{r69} achieves high-fidelity, clipping-free stabilization through Gaussian-based 3D scene reconstruction. Despite these advances, deep learning methods are often complex and resource-intensive. As an efficient alternative, our method adopts a hybrid strategy that integrates multi-threaded buffering technology into a three-stage processing pipeline.

\subsection{Feature Detection}

Classical detectors~\cite{r17,r18,r19} remain widely used, whereas deep methods~\cite{r29,r30,r31,r32,r33,r34,r35}
have been shown to improve robustness to illumination and viewpoint changes. In addition, methods~\cite{r36,r37} have achieved strong performance in low-texture and large-parallax cases. Existing stabilizers often rigidly couple detection, limiting adaptability. We propose a decoupled framework with keypoint uniformization and multi-detector collaboration, enabling scene-adaptive feature selection while retaining real-time efficiency.

\subsection{Optical Flow Estimation}

Variational methods~\cite{r38} pioneered the use of global optimization, while deep methods such as those in~\cite{r39,r40,r41,r42,r43,r44,r45,r46} achieved state-of-the-art accuracy. NeuFlow~\cite{r47} targets edge devices. For stabilization, optical flow can aggregate local motion cues into global priors. MemFlow~\cite{r48} demonstrated real-time performance by incorporating historical memory. Optical flow better captures non-rigid motion than top-down keypoint matching.



\section{Method}
\label{chap:Method}

Our pipeline consists of three stages: (i) motion estimation, (ii) motion propagation, and (iii) motion compensation. All three stages are strictly causal (i.e., past-only) and never access future frames. Training is fully unsupervised (see the self-supervised objectives in Secs.~\ref{sec:propagation} and~\ref{sec:compensation}); inference utilizes a multi-threaded, asynchronous pipeline for enhanced efficiency. Due to space constraints, full details of our multi-process acceleration strategy are deferred to the supplementary material (Sec.~C).

\begin{figure*}[t]
 \centering
  \includegraphics[width=0.9\linewidth]{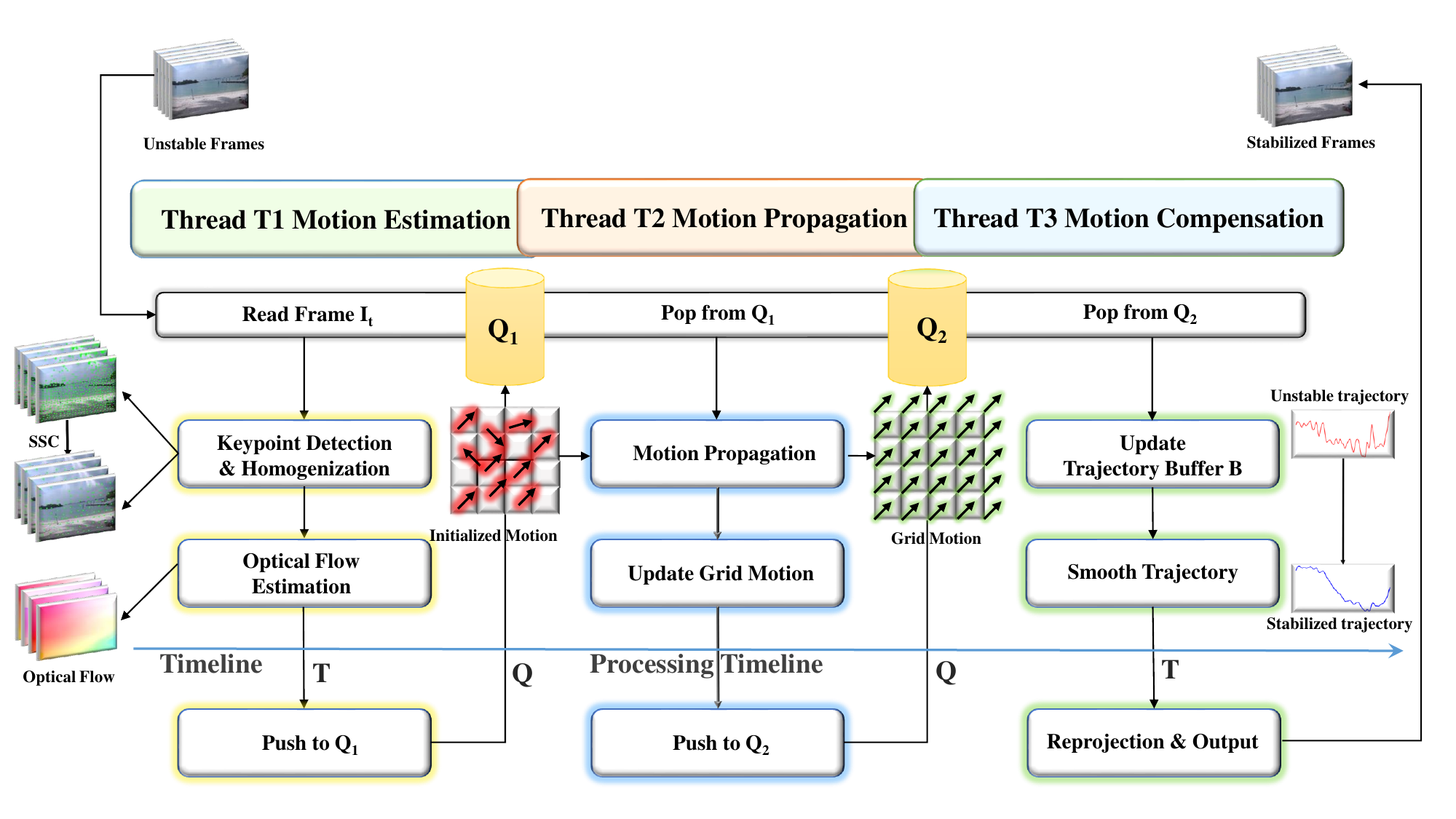}
  \caption{Overview of our video stabilization framework, consisting of three key modules: motion estimation, 
	motion propagation, and motion compensation. The motion estimation module detects keypoints and calculates 
	optical flow to estimate motion between frames. The motion propagation module transfers the estimated motion 
	information to a global trajectory buffer, ensuring motion consistency across frames. The motion compensation 
	module adjusts the frames based on the updated trajectories to produce a stabilized output. A multi-threaded 
	processing structure utilizes three threads, each dedicated to one of the core modules, and leverages shared 
	buffer queues for efficient parallel execution. This approach improves processing speed and online performance, 
	enabling the stabilization of video frames with minimal latency.}
  \label{fig:framework}
\end{figure*}

\subsection{Motion Estimation}
\label{sec:motion-estimation}
This section presents a strictly online observation pipeline that outputs keypoints $K_t$, displacements $\mathbf{u}_t$, and a causal dense backward flow $\hat{\mathbf{f}}_{t\leftarrow t-1}$. These serve as inputs to motion propagation and motion compensation. This step depends only on $\{I_{t-1}, I_t\}$ and never uses future frames.

\paragraph{Notation}
Let $I_t$ represent the $t$-th frame. We generate online: 
(i) candidate keypoints $\tilde K_t=\{(x_i, y_i, s_i)\}_i$ (with confidence $s_i$); (ii) the uniformized subset $K_t$; and (iii) a causal dense backward optical flow $\hat{\mathbf{f}}_{t\leftarrow t-1} \in \mathbb{R}^{H\times W\times 2}$, estimated from $\{I_{t-1}, I_t\}$. Next, we sample the observed displacements $\mathbf{u}_t$ at the keypoint locations and construct the feature representation for subsequent modules.

\paragraph{(1) Keypoint Collaboration Detection}
\label{sec:kpc}  
To reduce the computational cost and mitigate dynamic-object sensitivity in dense flow methods~\cite{r52, r53}, we adopt a flexible framework that incorporates feature points from heterogeneous detectors. Let  
\begin{equation}
	\label{eq:kp_detection}  
	\mathcal{D}=\big\{D^{\mathrm{trad}}_m\big\}_{m=1}^{M}  
	\cup\big\{D^{\mathrm{deep}}_n\big\}_{n=1}^{N},  
\end{equation} 
with each producing $\tilde K_t^{(j)}=\{(x_i,y_i,s_i)^{(j)}\}_i$. We normalize the confidence scores, apply NMS, and fuse the proposals from different sources using weighted averages: 
\begin{equation} 
	 \label{eq:kp_nms} 
	\tilde K_t=\operatorname{NMS}\!\Big(\bigcup_j w_j\cdot \tilde K_t^{(j)}\Big),  
\end{equation}  
where $w_j$ can be configured online according to scene characteristics or runtime budget.

\paragraph{(2) Homogenization}  
\label{sec:ssc}
To prevent clustering of keypoints in texture-rich regions, we employ Spatially Selective Clustering (SSC)~\cite{r54}. The image is partitioned into a $G_x\times G_y$ grid; for each cell $\mathcal{C}$, the top-$k$ keypoints are selected based on confidence $s$, subject to a minimum separation threshold $\tau$:  
\begin{equation}
	\label{eq:homogenization}
	\begin{split}
	K_t = \bigcup_{\mathcal{C}} \operatorname*{arg\,top}_k \Big\{ (x,y,s) \in \tilde K_t \cap \mathcal{C} \;\big| \\
	\min_{(x',y') \in K_t \cap \mathcal{C}} \|(x,y) - (x',y')\|_2 \ge \tau \Big\}.
	\end{split}
\end{equation}

\paragraph{(3) Sparse Keypoint-Guided Causal Flow Fusion}  

We estimate the causal dense backward flow $\mathbf{f}_{t\leftarrow t-1}$ using MemFlow~\cite{r48}.
To incorporate guidance from sparse keypoints, we build the mask using the denser candidate set $\tilde K_t$
(to improve coverage), while the uniformized subset $K_t$ is used for subsequent geometric estimation.

\begin{equation}
\label{eq:kp_sparse_mask}
M_t(x)=\mathbf{1}\!\left[\min_{p\in \tilde K_t}\|x-p\|_2\le r\right].
\end{equation}

where the mask is set to 1 within a circular neighborhood of radius $r$ centered at each keypoint, and 0 elsewhere.  

We then synthesize a reweighted flow field by directly using the estimated dense flow within the keypoint neighborhood, while using keypoint-based interpolated flow outside the neighborhood:  
\begin{equation}
	\label{eq:sparse_flow}
	\begin{split}
	\hat{\mathbf{f}}_{t\leftarrow t-1}(x) 
	&= M_t(x) \odot \mathbf{f}_{t\leftarrow t-1}(x) \\
	&\quad + (1 - M_t(x)) \odot 
	\mathcal{I}\!\big(\mathbf{f}_{t\leftarrow t-1}, \tilde K_t\big),
	\end{split}
\end{equation}

Finally, we extract the optical flow displacements at the keypoint locations from the reweighted flow field 
$\hat{\mathbf{f}}_{t\leftarrow t-1}$ as the observation:  
\begin{equation} 
	\label{eq:flow_displacements} 
	\mathbf{u}_t = \hat{\mathbf{f}}_{t\leftarrow t-1}(K_t)  
	\in \mathbb{R}^{|K_t| \times 2},  
\end{equation}  

and concatenate them with the keypoint coordinates to form the motion feature vector for subsequent tasks:  
\begin{equation}  
	\label{eq:motion_vector}
	\mathbf{m}_t = [x_{kp};\, y_{kp};\, u;\, v].  
\end{equation}

\subsection{Motion Propagation}
\label{sec:propagation}

While sparse keypoints and their associated initialized flow can capture local motion, dynamic and non-rigid scenes still pose challenges to modeling the camera’s global trajectory. To address this, we propose \textbf{EfficientMotionPro}, which anchors displacements to a regular grid using multi-homography priors, while focusing on learning only the non-rigid/parallax residuals. This design produces a full-frame grid motion field $\Delta g_t$. Importantly, both training and inference rely solely on past information (without access to future frames), and the entire process is conducted in a fully self-supervised manner. We denote the set of \emph{causal keypoints} at time $t$ as $\Omega_{kp}$ and the set of grid vertices as $\Omega_{grid}$.

\paragraph{(1) Multi-Grid Homography Prior}
\label{sec:homo}
Let $K_{\text{homo}}\!\ge\!1$ be a hyperparameter (the number of homography clusters). We cluster $\mathbf{u}_t$ into $K_{\text{homo}}$ groups via K-means and estimate a homography for each cluster with RANSAC, yielding $H_{t,1},\ldots,H_{t,K_{\text{homo}}}$. Let $g$ denote the regular grid vertices ($g\in\Omega_{grid}$); projecting $g$ produces $\{\hat g_{t,k}\}_{k=1}^{K_{\text{homo}}}$. We then compute soft fusion weights $\{\alpha_{t,k}(g)\}_{k=1}^{K_{\text{homo}}}$ based on local cluster prevalence or distance, with $\sum_{k}\alpha_{t,k}(g)=1$:
\begin{equation}
	\label{eq:base_motion}
	\hat g_t=\sum_{k=1}^{K_{\text{homo}}}\alpha_{t,k}(g)\,\hat g_{t,k},\qquad
	\Delta g_{\text{base},t}=g-\hat g_t.
\end{equation}

\paragraph{(2) Residual Network}
Given $\mathbf{m}_t=[x_{kp};\,y_{kp};\,u;\,v]$, a lightweight backbone (Ghost~\cite{r55}+ECA~\cite{r56}) predicts per-vertex residuals $\Delta g_{\text{res},t}$, from which we obtain
\begin{equation}
	\label{eq:Residual}
	\Delta g_t=\Delta g_{\text{base},t}+\Delta g_{\text{res},t}.
\end{equation}
All temporal operators are implemented as causal convolutions (left-padded).

\paragraph{(3) Core Self-Supervised Loss}
Our unsupervised objective enforces that the predicted grid displacements remain consistent with the observed causal keypoint motions. To improve robustness against noise and outliers, we adopt the Charbonnier penalty~\cite{r63} and further introduce per-keypoint confidence weights $\omega_{t,p}\in[0,1]$ that adaptively downweight unreliable keypoints:
\begin{equation}
	\label{eq:kp_loss}
	\mathcal{L}_{\text{kp}} = \tfrac{1}{|\Omega_{kp}|}\sum_{p\in\Omega_{kp}} 
	\omega_{t,p}\,\sqrt{\lVert \mathbf{u}_{t,p}-\Delta g_t(p)\rVert_2^2 + \epsilon^2},
\end{equation}
where $\epsilon$ is a small stabilizing constant (e.g., $10^{-3}$).

This confidence-weighted keypoint consistency loss serves as the core driver of propagation. Additional regularization terms (e.g., projection consistency and a structure-preservation constraint) are incorporated to further improve stability and prevent degenerate solutions; see supplement (Sec.~A).

\subsection{Motion Compensation}
\label{sec:compensation}

In dynamic scenes or under low-frame-rate conditions, grid vertex trajectories often exhibit local oscillations due to jitter, occlusion, or non-rigid disturbances, which compromise the stability of the overall motion field. To mitigate this issue, we introduce \textbf{OnlineSmoother}, a lightweight online trajectory smoothing module inspired by~\cite{r15,r25}. Without relying on future frames, it employs a learnable causal kernel to smooth propagation-generated trajectories, followed by frame-wise grid deformation that produces stabilized videos. Neither training nor inference requires annotations or ground-truth data.

\paragraph{(1) Causal Kernel Smoother}
\label{sec:smoother}
A left-padded causal convolution encodes long-term memory; the decoder predicts $3$-tap causal kernels for the $x$ and $y$ directions ($6$ channels total), corresponding to an effective temporal window of $L{=}7$ frames. We denote the kernel tensor by
\begin{equation}
	\label{eq:kernel_concat}
	\mathcal{K}=\mathrm{concat}(\mathcal{K}^x,\mathcal{K}^y)\in\mathbb{R}^{B\times 6\times T\times H\times W},
\end{equation}
where $B$ is the batchsize, $T$ is the number of causal time steps held in the buffer, and $H{\times}W$ is the grid resolution (number of vertices in $\Omega_{grid}$). For the $x$ direction at time $t$,
\begin{equation}
	\label{eq:causal_smoother}
	S_t^x =
	\frac{
	\lambda \sum_{r\in\{1,2,3\}} k_{t,r}^x \, S_{t-r}^x
	+ O_t^x
	}{
	1 + \lambda \sum_{r} |k_{t,r}^x|
	},
\end{equation}
where $O_t$ is the input trajectory (e.g., grid-vertex trajectories from $\Delta g_t$ or its integral) and $\lambda{=}100$ by default. Only cached past states are accessed (no look-ahead). We empirically set $L{=}7$ and use $\Delta_{\max}{=}\lfloor(L-1)/2\rfloor{=}3$ in the temporal loss below; see figure~\ref{fig:ablation} for ablations.

\paragraph{(2) Core Self-Supervised Loss (definitions included)}
To ensure stable yet responsive trajectories under causal constraints, we combine a time-adaptive second-order penalty with a frequency prior. Let $p\in\Omega_{grid}$ index grid vertices. The adaptive temporal term is
\begin{equation}
	\label{eq:time_loss}
	\begin{split}
	\mathcal{L}_{\text{time}}
	&=\tfrac{1}{|\Omega_{grid}|}\sum_{p\in\Omega_{grid}}\sum_{\Delta=1}^{\Delta_{\max}}
	\frac{\alpha_\Delta\,e^{-\beta\|S_t(p)-S_{t-\Delta}(p)\|_2^2}}{\Delta^2} \\
	&\quad\cdot\sqrt{\|S_t(p)-2S_{t-\Delta}(p)+S_{t-2\Delta}(p)\|_2^2+\epsilon^2}.
	\end{split}
\end{equation}
\textbf{Definitions.}
(i) $\Delta_{\max}\in\mathbb{N}$ is the causal horizon used by the loss (we set $\Delta_{\max}{=}3$ for $L{=}7$ so that the largest stencil $t-2\Delta$ remains within the causal window).  
(ii) $\alpha_\Delta\!\ge\!0$ are fixed, distance-decaying weights
\begin{equation}
\label{eq:time_weight}
\alpha_\Delta
=\frac{\exp(-\Delta/\tau_{\text{time}})}{\sum_{d=1}^{\Delta_{\max}}\exp(-d/\tau_{\text{time}})},
\end{equation}
\noindent where $\tau_{\text{time}}=2$ by default.

(iii) $\beta\!\ge\!0$ controls motion-adaptive attenuation; we use a global \emph{learnable} scalar with softplus reparameterization, initialized to $\beta_0{=}0.02$. Larger $\beta$ downweights the penalty when recent motion magnitude $\|S_t{-}S_{t-\Delta}\|$ is large.

The frequency term suppresses high-frequency oscillations within each causal window of length $L$:
\begin{equation}
	\label{eq:freq_loss}
	\mathcal{L}_{\text{freq}}
	=\tfrac{1}{|\Omega_{grid}|}\sum_{p\in\Omega_{grid}}\sum_{m=1}^{\lfloor L/2\rfloor}
	\gamma(\omega_m)\,\|\hat S_p(\omega_m)\|_2^2,
\end{equation}
where $\hat S_p(\omega_m)$ is the DFT magnitude of $\{S_{t-d}(p)\}_{d=0}^{L-1}$ at frequency $\omega_m{=}2\pi m/L$, and the frequency weights are \(\gamma(\omega_m)=\gamma_0\!\left(\frac{\omega_m}{\omega_N}\right)^{2}\) with \(\omega_N=\pi\) and \(\gamma_0\ge 0\), by default we set \(\gamma_0=0.1\).

The overall temporal objective is
\begin{equation}
	\label{eq:temp_loss}
	\mathcal{L}_{\text{temp}}
	= \lambda_{\text{time}} \mathcal{L}_{\text{time}}
	+ \lambda_{\text{freq}} \mathcal{L}_{\text{freq}},
\end{equation}
with $\lambda_{\text{time}},\lambda_{\text{freq}}\!>\!0$ (default $\lambda_{\text{time}}  = 1.0$, $\lambda_{\text{freq}} = 0.1$). All loss terms are fully causal.

\paragraph{(3) Frame-Level Compensation and Rendering}
Let $M_t = S_t - O_t$ denote the compensation displacement field (defined on grid vertices). For each pixel $x$, obtain $M_t(x)$ by bilinear interpolation from vertex displacements, and perform either forward grid warping or backward sampling:
\begin{equation}
	\label{eq:frame_warping}
	x' = x + M_t(x),\qquad
	\tilde I_t(x) = I_t\!\big(\mathcal{W}(x;M_t)\big).
\end{equation}
where $\mathcal{W}$ is the backward mapping with bilinear resampling to avoid holes. We employ ProPainter~\cite{r71} for outpainting at frame boundaries to minimize black borders; see supplement (Sec.~D) for details.

\subsection{UAV-Test Dataset}

Most existing video stabilization benchmarks are collected with handheld devices under visible-light conditions, which limits their applicability in real-world scenarios. In contrast, unstable videos frequently arise in safety-critical domains such as UAV-based night-time remote sensing and guided munitions. To bridge this gap, we introduce the \textbf{UAV-Test} dataset, comprising unstable aerial videos in both visible and infrared modalities. \textbf{UAV-Test} contains 92 sequences spanning five representative scenarios: urban areas, highways, forests/mountains, waterfronts, and industrial sites. Summary statistics are reported in Table~\ref{tab:uavtest}, with additional implementation details and extended dataset analysis provided in the supplemental material (Sec.~E).

\begin{table*}[t]
\centering
\caption{Overview of the UAV-Test dataset.}
\resizebox{\textwidth}{!}{
\begin{tabular}{l c c c c c c}
\hline
\textbf{Scene Type} & \textbf{Sensor} & \textbf{Tilt Range} & \textbf{Resolution} & \textbf{FPS} & \textbf{Occlusion} & \textbf{Sequences}\\
\hline
Urban areas       & Visible/IR & $30^\circ$- $90^\circ$ & $1920 \times 1080$ & 25 & Partial building occlusion & 19 \\
Highways          & Visible/IR & $30^\circ$- $90^\circ$ & $1280 \times 720$  & 30 & Minimal occlusion          & 22 \\
Forests/mountains & Visible/IR & $45^\circ$- $90^\circ$ & $1920 \times 1080$ & 25 & Frequent canopy occlusion  & 18 \\
Waterfronts       & Visible    & $60^\circ$- $90^\circ$ & $1920 \times 1080$ & 30 & Mild vapor occlusion       & 20 \\
Industrial sites  & Visible/IR & $30^\circ$- $60^\circ$ & $1280 \times 720$  & 25 & Large machinery occlusion  & 13 \\
\hline
\end{tabular}
}
\label{tab:uavtest}
\end{table*}

\section{Experiment}
\label{chap:experiment}

\subsection{Quantitative Evaluation}

\begin{table*}[t]
\centering
\caption{Quantitative results on the NUS~\cite{r20}, DeepStab~\cite{r10}, Selfie~\cite{r11}, GyRo~\cite{r68} and UAV-Test datasets. We report Cropping Ratio (C), Distortion Value (D), and Stability Score (S). All metrics range from 0 to 1; higher is better. Best results in each group (offline or online) are in \textcolor{red}{red}, second best in \textcolor{blue}{blue}. A runtime (fps) comparison among online methods is provided in the supplementary material (Sec.~F).}
\label{tab:quantitative}
\setlength{\tabcolsep}{2pt}
\begin{tabular}{lc ccc ccc ccc ccc ccc}
\toprule
\textbf{} & \multirow{2}{*}{\textbf{Method}} 
& \multicolumn{3}{c}{\textbf{NUS}}
& \multicolumn{3}{c}{\textbf{DeepStab}}
& \multicolumn{3}{c}{\textbf{Selfie}} 
& \multicolumn{3}{c}{\textbf{GyRo}}
& \multicolumn{3}{c}{\textbf{UAV-Test}}\\
\cmidrule(lr){3-5} \cmidrule(lr){6-8} \cmidrule(lr){9-11} \cmidrule(lr){12-14} \cmidrule(lr){15-17}
& & C↑ & D↑ & S↑ & C↑ & D↑ & S↑ & C↑ & D↑ & S↑ & C↑ & D↑ & S↑ & C↑ & D↑ & S↑\\
\midrule
\multirow{13}{*}{\textbf{Offline}} 
& Grundmann et al.~\cite{r1} & 0.71 & 0.76 & 0.82 & 0.77 & 0.87 & 0.84 & 0.75 & 0.81 & 0.83 & 0.91 & 0.89 & 0.81 & 0.72 & 0.64 & 0.76 \\
& Bundle~\cite{r20} & 0.81 & 0.78 & 0.82 & 0.80 & 0.90 & 0.85 & 0.74 & 0.82 & 0.80 & 0.93 & 0.82 & 0.85 & 0.79 & 0.69 & 0.88 \\
& DIFRINT~\cite{r24} & \textcolor{red}{1.00} & 0.87 & 0.84 & \textcolor{red}{1.00} & 0.91 & 0.78 & \textcolor{red}{1.00} & 0.78 & 0.84 & \textcolor{red}{1.00} & 0.83 & 0.85 & \textcolor{red}{1.00} & 0.86 & 0.87 \\
& Yu and Ramamoorthi~\cite{r4} & 0.85 & 0.81 & 0.86 & 0.87 & 0.92 & 0.82 & 0.83 & 0.79 & 0.86 & \textcolor{red}{1.00} & 0.92 & 0.84 & 0.85 & 0.83 & 0.86 \\
& PWStableNet~\cite{r21}  & 0.91 & 0.97 & 0.86 & 0.97 & 0.97 & 0.87 & \textcolor{blue}{0.99} & 0.92 & 0.89 & \textcolor{blue}{0.99} & 0.96 & 0.83 & 0.90 & 0.88 & 0.90\\
& DUT~\cite{r13} & \textcolor{blue}{0.98} & 0.88 & 0.85 & \textcolor{blue}{0.99} & 0.95 & \textcolor{blue}{0.95} & \textcolor{blue}{0.99} & \textcolor{red}{0.98} & \textcolor{blue}{0.93} & \textcolor{blue}{0.99} & \textcolor{blue}{0.98} & 0.89 & \textcolor{blue}{0.95} & 0.89 & \textcolor{red}{0.94} \\
& Deep3D~\cite{r7} & 0.66 & 0.90 & \textcolor{red}{0.94} & 0.75 & \textcolor{blue}{0.98} & 0.92 & 0.35 & 0.70 & \textcolor{red}{0.95} & \textcolor{blue}{0.99} & 0.94 & 0.77 & 0.68 & 0.89 & 0.86 \\
& FuSta~\cite{r27} & \textcolor{red}{1.00} & 0.87 & 0.86 & \textcolor{red}{1.00} & 0.83 & 0.87 & \textcolor{red}{1.00} & 0.92 & 0.82 & \textcolor{red}{1.00} & 0.97 & 0.87 & \textcolor{red}{1.00} & 0.88 & 0.83 \\
& RStab~\cite{r28} & \textcolor{red}{1.00} & \textcolor{red}{0.99} & \textcolor{red}{0.94} & \textcolor{red}{1.00} & \textcolor{blue}{0.98} & \textcolor{red}{0.96} & \textcolor{red}{1.00} & 0.92 & \textcolor{red}{0.95} & \textcolor{red}{1.00} & 0.95 & \textcolor{blue}{0.92} & \textcolor{red}{1.00} & \textcolor{blue}{0.96} & \textcolor{red}{0.94} \\
& MetaStab~\cite{r67} & \textcolor{red}{1.00} & \textcolor{red}{0.99} & 0.91 & \textcolor{red}{1.00} & \textcolor{red}{0.99} & 0.91 & \textcolor{red}{1.00} & \textcolor{blue}{0.96} & 0.58 & \textcolor{red}{1.00} & \textcolor{red}{0.99} & 0.90 & \textcolor{red}{1.00} & \textcolor{red}{0.99} & \textcolor{blue}{0.91} \\
& VidStabFormer~\cite{r64} & \textcolor{red}{1.00} & 0.90 & 0.85 & \textcolor{red}{1.00} & 0.90 & 0.82 & \textcolor{red}{1.00} & 0.85 & 0.88 & \textcolor{red}{1.00} & 0.89 & 0.86 & \textcolor{red}{1.00} & 0.87 & 0.84 \\
& Gavs~\cite{r69} & \textcolor{red}{1.00} & \textcolor{blue}{0.98} & \textcolor{blue}{0.93} & \textcolor{red}{1.00} & 0.96 & 0.90 & \textcolor{red}{1.00} & 0.78 & 0.85 & \textcolor{red}{1.00} & \textcolor{red}{0.99} & \textcolor{red}{0.93} & \textcolor{red}{1.00} & 0.83 & 0.87 \\
& Adobe Premiere Pro (2025) & 0.84 & 0.82 & 0.91 & 0.73 & 0.87 & 0.87 & 0.71 & 0.80 & 0.84 & 0.73 & 0.96 & 0.60 & 0.75 & 0.86 & 0.90 \\
\midrule
\multirow{5}{*}{\textbf{Online}} 
& MeshFlow~\cite{r23} & 0.78 & 0.83 & 0.85 & 0.81 & 0.87 & \textcolor{red}{0.86} & 0.75 & 0.88 & 0.84 & \textcolor{blue}{0.79} & 0.88 & 0.82 & 0.77 & 0.83 & 0.70 \\
& StabNet~\cite{r10} & 0.66 & 0.88 & 0.82 & 0.65 & 0.86 & 0.80 & 0.70 & 0.78 & 0.83 & 0.69 & 0.85 & 0.83 & 0.73 & 0.84 & 0.73 \\
& NNDVS~\cite{r15} & \textcolor{blue}{0.92} & \textcolor{red}{0.98} & 0.87 & \textcolor{blue}{0.93} & \textcolor{red}{0.91} & 0.84 & \textcolor{blue}{0.97} & \textcolor{blue}{0.92} & \textcolor{red}{0.91} & \textcolor{red}{0.99} & 0.93 & 0.88 & \textcolor{blue}{0.89} & 0.87 & 0.84 \\
& Liu et al.~\cite{r22} & 0.72 & \textcolor{blue}{0.89} & \textcolor{blue}{0.89} & 0.89 & \textcolor{blue}{0.88} & \textcolor{blue}{0.85} & 0.79 & 0.89  & \textcolor{blue}{0.85} & \textcolor{red}{0.99} & \textcolor{blue}{0.94} & \textcolor{blue}{0.89} & 0.82 & \textcolor{blue}{0.89} & \textcolor{blue}{0.85} \\
& \textbf{Ours} & \textcolor{red}{0.95} & \textcolor{red}{0.98} & \textcolor{red}{0.90} & \textcolor{red}{0.94} & \textcolor{red}{0.91} & \textcolor{blue}{0.85} & \textcolor{red}{0.98} & \textcolor{red}{0.93} & \textcolor{red}{0.91} & \textcolor{red}{0.99} & \textcolor{red}{0.96} & \textcolor{red}{0.93} & \textcolor{red}{0.94} & \textcolor{red}{0.90} & \textcolor{red}{0.89} \\
\bottomrule
\end{tabular}
\end{table*}

\noindent\textbf{Baselines.} 
We evaluate our method against a diverse set of representative stabilization algorithms, covering both offline and online approaches. Offline baselines include~\cite{r1,r20,r24,r4,r21,r13,r7,r27,r28,r67,r64,r69} together with the commercial stabilizer \textit{Adobe Premiere Pro 2025}. Online methods include~\cite{r23,r10,r15,r22}. For comparison, we use either official videos or results from official implementations with default settings or pretrained models.

\noindent\textbf{Datasets.} 
We evaluate on three public benchmarks and an additional multimodal UAV dataset:  
(1) \textit{NUS}~\cite{r20}, 144 sequences across six categories (Regular, Running, Zooming, Crowd, Parallax, QuickRotating);  
(2) \textit{DeepStab}~\cite{r10}, 60 pairs of synchronized unstable/stable videos;  
(3) \textit{Selfie}~\cite{r11}, 33 front camera sequences with severe shaking;  
(4) \textit{GyRo}~\cite{r68}, which consists of 50 videos with precise alignment of gyroscope and OIS data; the test set is divided into six categories: GENERAL, ROTATION, PARALLAX, DRIVING, PEOPLE, and RUNNING;  
(5) \textit{UAV-Test}, consisting of 92 videos (74 RGB, 18 infrared) captured by UAVs across diverse scenarios, including urban highways, forests, dense cities, industrial sites, and waterfronts.  

\noindent\textbf{Metrics.} 
Following prior works~\cite{r13,r7,r27,r28,r10,r15}, we report three metrics:  
(1) \textit{Cropping Ratio (C)}: content preservation after stabilization (higher is better);  
(2) \textit{Distortion Value (D)}: anisotropy of estimated homographies, indicating geometric fidelity (higher is better);  
(3) \textit{Stability Score (S)}: low-frequency energy ratio, reflecting motion smoothness (higher is better).  
Together, these metrics capture content integrity, geometric fidelity, and temporal stability.

\noindent\textbf{Results on the NUS dataset.}
On the NUS benchmark~\cite{r20} (Table~\ref{tab:quantitative}), our method achieves the best online performance across all three metrics ($C=0.95$, $D=0.98$, $S=0.90$). Compared with NNDVS~\cite{r15}, our approach matches its geometric fidelity ($D=0.98$) while surpassing it in stability and cropping. This highlights our advantage in maintaining boundary completeness and suppressing distortion, while remaining close to the strongest offline method RStab~\cite{r28}.

\noindent\textbf{Results on the DeepStab dataset.}
For DeepStab~\cite{r10} (paired 720p), our method ranks first among online methods in Cropping Ratio ($C=0.94$), ties with NNDVS in geometric fidelity ($D=0.91$), and achieves competitive stability ($S=0.85$, second only to MeshFlow~\cite{r23} at $0.86$). This shows good transfer from low resolution, fast shaking scenarios to higher definition inputs, achieving a balanced trade-off between smoothness and content preservation.

\noindent\textbf{Results on the Selfie dataset.}
The Selfie dataset~\cite{r11}, with severe handheld shake and large motion, poses challenges to stabilization. As reported in Table~\ref{tab:quantitative} (center right), our method delivers the best online performance with $C=0.98$, $D=0.93$, and $S=0.91$. This confirms that our model handles extreme motion and dynamic scenes better than traditional geometric and deep content-driven baselines.

\noindent\textbf{Results on the GyRo dataset.}
On the GyRo benchmark~\cite{r68}, which involves gyroscope-based scenarios, our method achieves state-of-the-art online performance with Cropping Ratio ($C=0.99$), Distortion Value ($D=0.96$), and Stability Score ($S=0.93$). This outperforms other online approaches, such as NNDVS~\cite{r15} ($C=0.99$, $D=0.93$, $S=0.88$) and Liu et al.~\cite{r22} ($C=0.99$, $D=0.94$, $S=0.89$), and is competitive with top offline methods like Gavs~\cite{r69} ($C=1.00$, $D=0.99$, $S=0.93$), demonstrating the effectiveness of our causal framework.

\noindent\textbf{Results on the UAV-Test dataset.}
Finally, on the new UAV-Test benchmark with visible/IR sequences across diverse scenarios, our method sets a new online state of the art: $C=0.94$, $D=0.90$, $S=0.89$, outperforming NNDVS~\cite{r15} ($C=0.89$, $D=0.87$, $S=0.84$) and Liu et al.~\cite{r22} ($C=0.82$, $D=0.89$, $S=0.85$). Compared with the strongest offline models (e.g., RStab~\cite{r28}), our results remain competitive while relying only on causal information, showcasing strong generalization across modalities and complex UAV scenarios.


\subsection{Qualitative Results} 
\begin{figure}[t]
 \centering
  \includegraphics[width=0.99\linewidth]{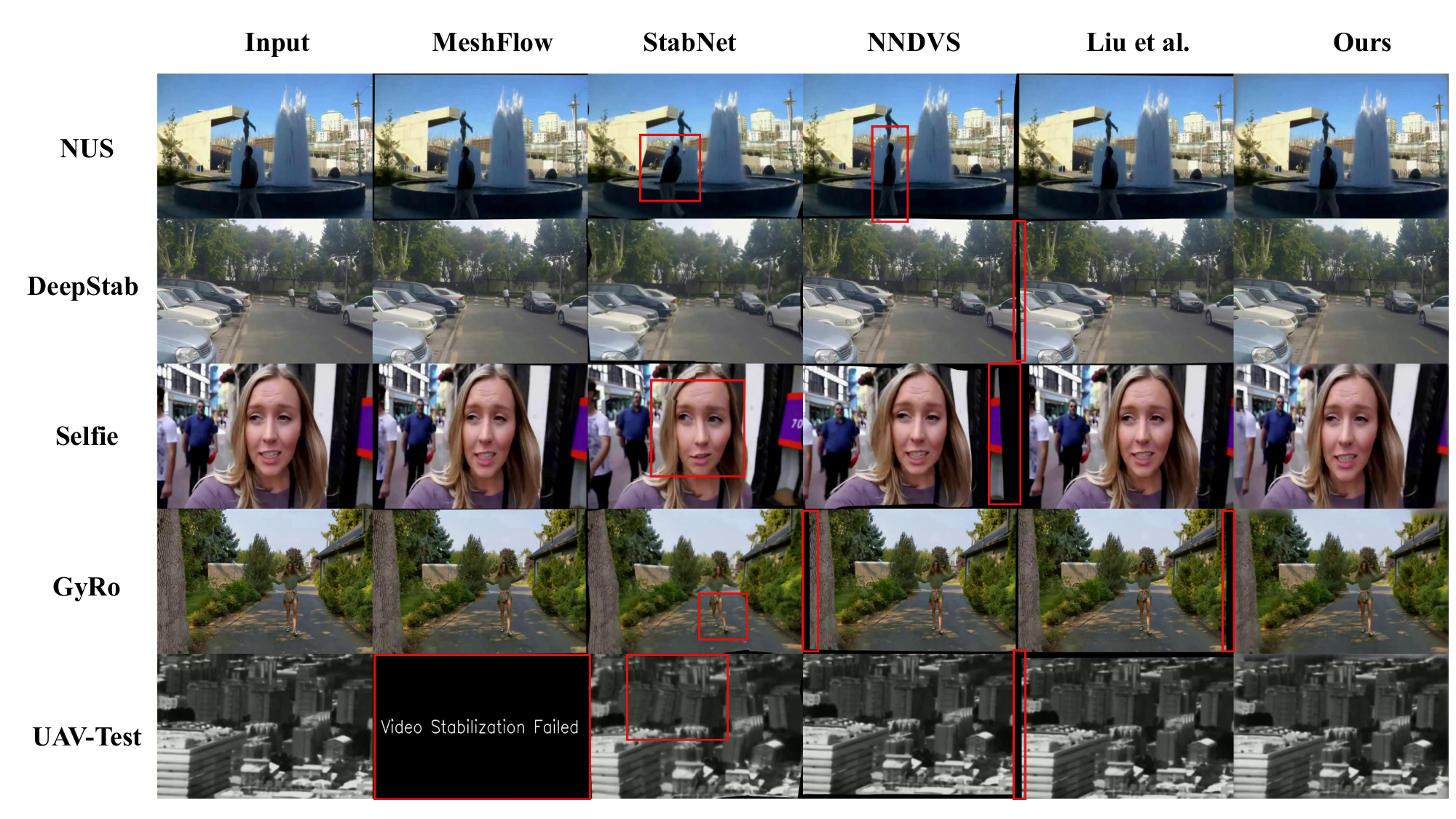}
  \caption{Comparison of visual results across different online methods. Distorted regions and black borders are 
  highlighted with red boxes.}
  \label{fig:visualizationperscene}
\end{figure}

The visual comparison results in Figure~\ref{fig:visualizationperscene} show that competing online methods~\cite{r23,r10,r15,r22} often produce artifacts such as shear, distortion, or excessive cropping. In contrast, our method generates fewer artifacts across various scenes and is more effective at preserving scene structure, demonstrating superior generalization without the use of future frames.

\subsection{User Study}
We conducted a user study to compare our method with other online approaches~\cite{r23,r10,r15,r22}. Five datasets were prepared, each containing two randomly selected video types. Fifty participants ranked the videos from best to worst based on their preference. The videos were randomly ordered and included the original unstable videos. The results, shown in Figure~\ref{fig:user_study}, indicate that users generally preferred our videos as the best. See the supplemental material (Sec.~I) for details.

\begin{figure}[t]
 \centering
  \includegraphics[width=0.99\linewidth]{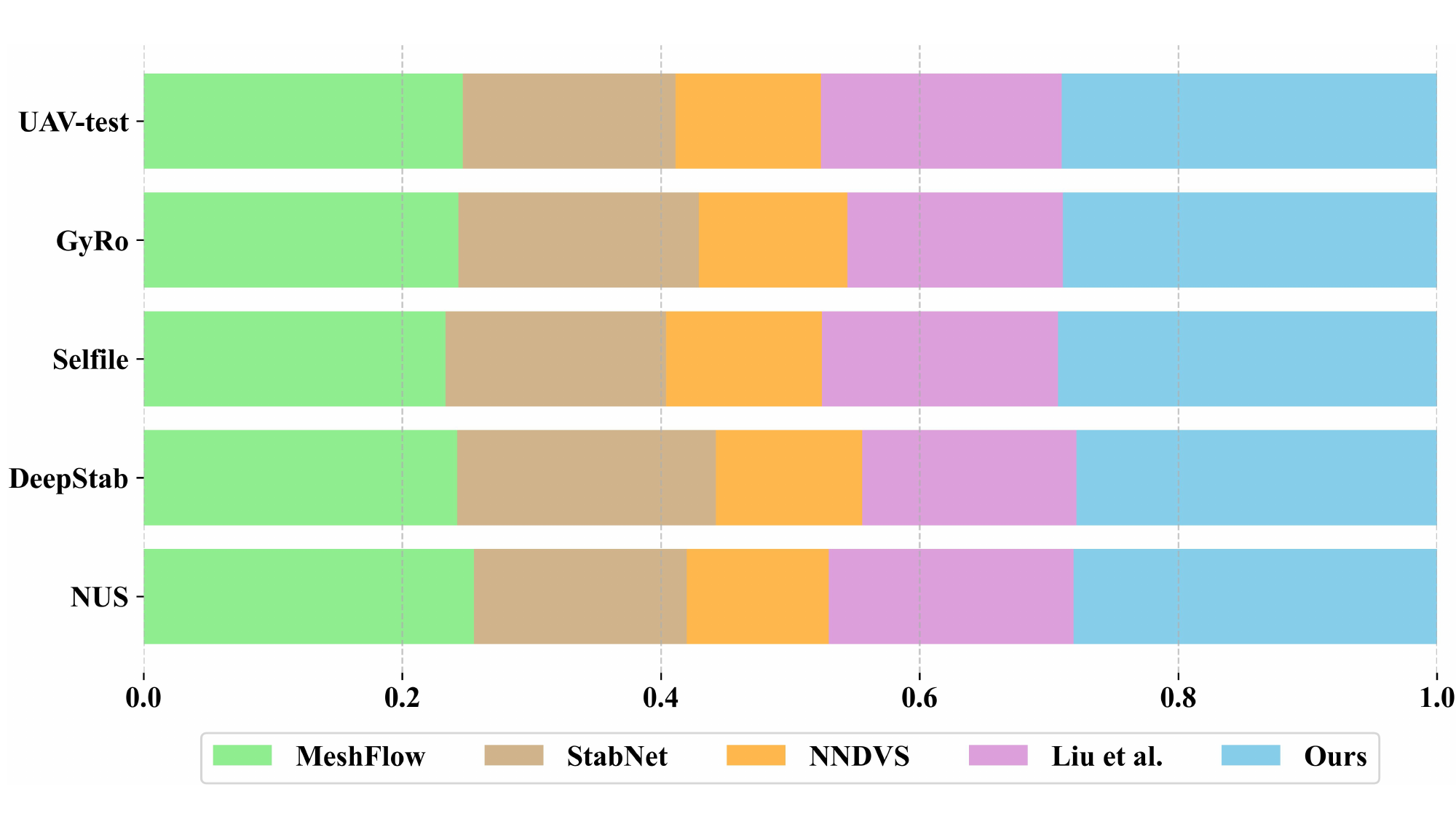}
  \caption{User study preferences.}
  \label{fig:user_study}
\end{figure}

\section{Ablation Study}
\label{chap:ablation}

\begin{figure}[t]
 \centering
  \includegraphics[width=0.99\linewidth]{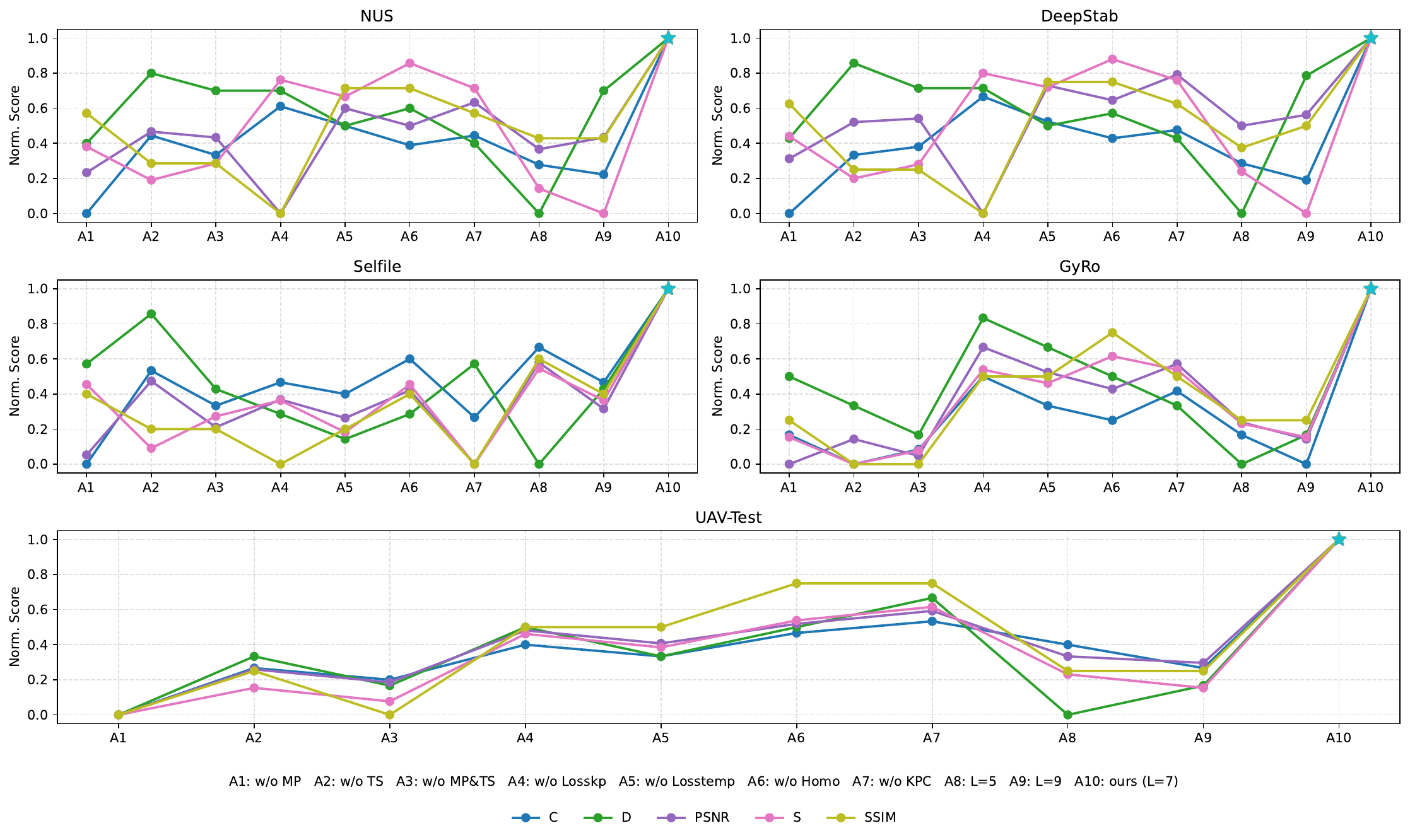}
  \caption{Normalized ablation study results on five datasets (NUS, DeepStab, Selfile, GyRo, UAV-Test). The x-axis denotes ablation configurations A1--A9 (variants with modules removed/modified) and A10 (our full model with window length \(L=7\)). The y-axis shows normalized scores (higher is better) for five metrics: Consistency (C), Distortion Value (D), Peak Signal-to-Noise Ratio (PSNR), Stability Score (S), and Structural Similarity (SSIM).}
  \label{fig:ablation}
\end{figure}

We perform ablation experiments on multiple datasets, as shown in Fig.~\ref{fig:ablation}, where A10 is the full model (\(L=7\)). The results demonstrate the effectiveness of each component:

\begin{itemize}
    \item \textbf{w/o MP (Sec.~\ref{sec:propagation}):} Removing motion propagation (A1) degrades D and PSNR, confirming its importance for global motion modeling.
    \item \textbf{w/o TS (Sec.~\ref{sec:smoother}):} Without trajectory smoothing (A2), structural stability decreases, as reflected in higher D.
    \item \textbf{w/o MP\&TS:} Disabling both (A3) causes the most severe drop, highlighting their complementarity.
    \item \textbf{w/o Loss\(_{kp}\) (Eq.~\ref{eq:kp_loss}):} Excluding keypoint consistency loss (A4) weakens motion supervision, lowering D and slightly PSNR.
    \item \textbf{w/o Loss\(_{temp}\) (Eq.~\ref{eq:temp_loss}):} Without temporal regularization (A5), grid trajectories become less stable, increasing D while PSNR remains similar.
    \item \textbf{w/o Homo (Sec.~\ref{sec:homo}):} Replacing multi-homography with a single homography (A6) introduces jitter and local distortions, reducing D and SSIM.
    \item \textbf{w/o KPC (Sec.~\ref{sec:motion-estimation}):} Using raw keypoints without collaboration (A7) leads to uneven sampling and lower D.
    \item \textbf{Window length (Sec.~\ref{sec:smoother}):} Shorter windows (A8, \(L=5\)) improve stability but reduce fidelity, while longer windows (A9, \(L=9\)) increase cost without consistent gains. \(L=7\) (A10) provides the best trade-off.
\end{itemize}

Overall, the full model achieves the highest scores, validating the contribution of each module.
\section{Conclusion}
\label{chap:Conclusion}
In this paper, we present a novel, lightweight, unsupervised framework for online video stabilization that overcomes the limitations of conventional end-to-end models, namely strong reliance on paired datasets, limited controllability, and difficulty of deployment on resource-constrained devices. The framework comprises three stages: motion estimation, motion propagation, and motion compensation, together with a system-level multithreaded acceleration strategy. By introducing collaborative keypoint detection and a keypoint uniformization scheme, the framework achieves accurate and robust motion estimation. Coupled with grid-based residual propagation and a dynamic smoothing kernel generation mechanism, it enables fine-grained modeling of complex motion trajectories with online smoothing. A multithreaded buffering mechanism further ensures stable operation on embedded platforms. In addition, we introduce a new multimodal aerial view dataset of unstable videos, UAV-Test, covering diverse geographic scenes and viewpoints to complement existing benchmarks. Extensive experiments show that our method outperforms state-of-the-art online approaches overall and approaches the performance of offline methods, demonstrating strong potential for real-time applications.


\section{Appendix}
The supplementary material additionally provides the following:
\begin{itemize}
\item Implementation details of the motion propagation networks;
\item Implementation details of the trajectory smoothing networks;
\item Details of the multi-threaded asynchronous buffering pipeline;
\item Details of frame outpainting;
\item Information on the multimodal UAV dataset;
\item Runtime comparison on embedded platforms;
\item Comprehensive description of the evaluation metrics used in this work;
\item Per-scene quantitative evaluation of online methods;
\item Additional visualizations of key modules and optical flow;
\item A 3-minute video demonstrating comparisons with online methods;
\item Limitations and future directions.
\end{itemize}

\section{Details of EfficientMotionPro}

\subsection{Training Objectives}  
We propose a grid-based motion propagation network (\textbf{EfficientMotionPro}). Building upon multi-homography priors, this module explicitly encodes the relationship between sparse keypoint residuals and vertex distances. Through lightweight feature extraction, attention-weighted aggregation, and residual decoding, it achieves end-to-end propagation from sparse keypoints to a dense grid, thereby yielding a more robust global motion field. We additionally employ projection consistency and structure-preservation objectives for training.

\paragraph{(1) Homography Projection Constraint}  
To enhance the geometric fidelity of the propagated motion field, we introduce a homography-based projection loss. 
Specifically, the estimated keypoint motion is projected via a local homography mapping generated from the predicted 
grid motion vectors. The Euclidean distance between this projected position and the actual ground-truth keypoint 
position after motion constitutes the projection error:
\begin{equation}
	\label{eq:proj_loss}
	\mathcal{L}_{\text{proj}} = \tfrac{1}{|\Omega_{kp}|}\sum_{p\in\Omega_{kp}} \omega_{t,p}\,
	\sqrt{ \left\lVert \mathbf{u}_{t,p} - 
	H_{t}^{\text{local}}(p;\,\Delta g_t) \cdot p \right\rVert_2^2 + \epsilon^2 }.
\end{equation}
Here, $H_{t}^{\text{local}}(p;\,\Delta g_t)$ denotes the local homography transformation, obtained via bilinear sampling from 
the grid motion field $\Delta g_t$ surrounding point $p$. This term effectively constrains the grid deformation 
predicted by the residual network to be geometrically plausible, reducing inconsistent distortions.

\paragraph{(2) Structure-Preservation Constraint}  
In low-texture regions, the propagation estimate is prone to local distortions or discontinuous boundaries. To maintain 
the rigid structure of the mesh, we design a structure-preserving loss based on the angular relationships between adjacent 
vertices. For each grid cell defined by its four vertices $\{g_{i,j}, g_{i+1,j}, g_{i,j+1}, g_{i+1,j+1}\}$, we enforce orthogonality between adjacent edges:
\begin{equation}
	\label{eq:struct_loss}
	\mathcal{L}_{\text{struct}} = \tfrac{1}{|G|} \sum_{i,j} \left(
	\frac{ \big\langle g_{i+1,j} - g_{i,j},\;
	R_{90^\circ}\!\big(g_{i,j+1} - g_{i,j}\big) \big\rangle }
	{ \left\lVert g_{i+1,j} - g_{i,j} \right\rVert
	\cdot \left\lVert g_{i,j+1} - g_{i,j} \right\rVert }
	\right)^2,
\end{equation}
where $R_{90^\circ}$ denotes a $90^\circ$ rotation operator, $\langle \cdot, \cdot \rangle$ is the dot product, and $|G|$ is 
the total number of grid cells. This loss penalizes deviations from orthogonality, suppressing shear and anisotropic 
scaling, thereby preserving local shape and preventing distortion.

\paragraph{(3) Total Loss}  
The final optimization objective for the motion-propagation module is defined as the weighted combination of all three 
losses:
\begin{equation}
	\label{eq:motionpro_total_loss}
	\mathcal{L} = \lambda_1 \mathcal{L}_{\text{kp}} + \lambda_2 \mathcal{L}_{\text{proj}} + \lambda_3 \mathcal{L}_{\text{struct}},
\end{equation}
with default settings $(\lambda_1,\lambda_2,\lambda_3)=(10,40,40)$.

\begin{figure}[t]
  \centering
  \includegraphics[width=0.45\textwidth]{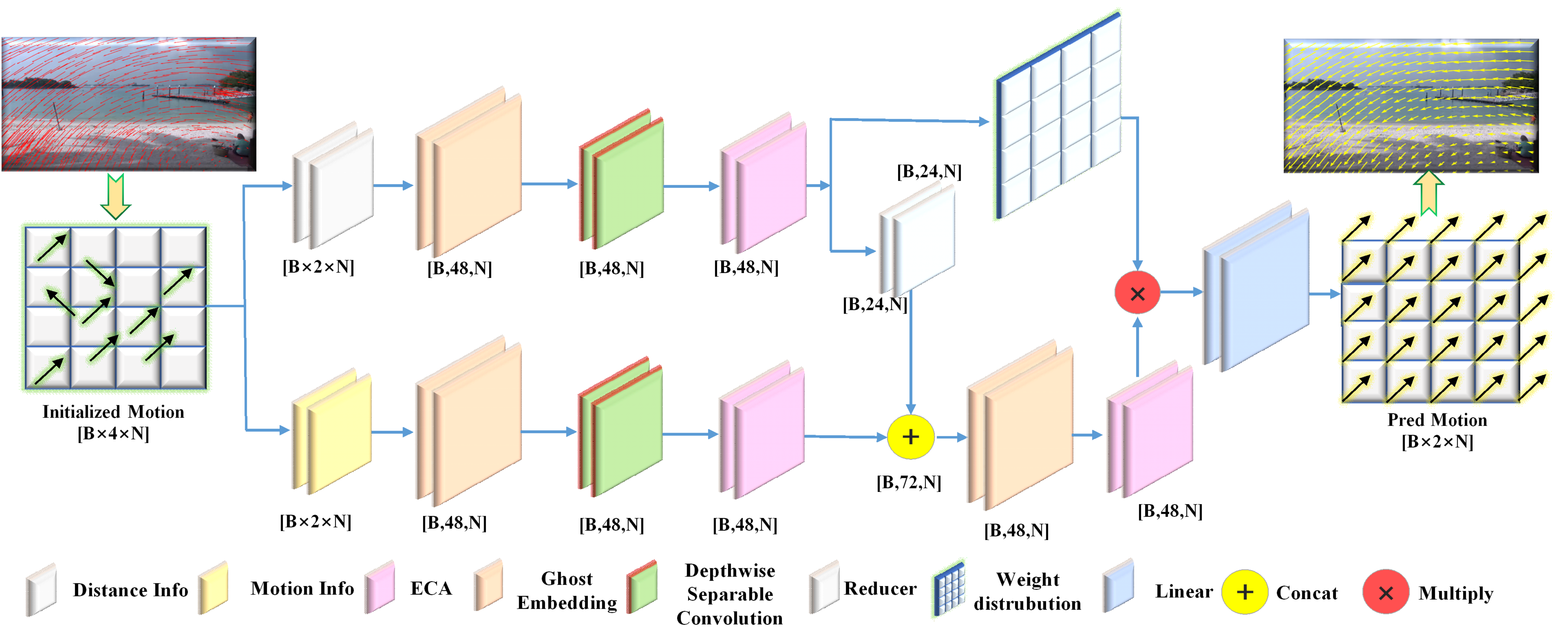}
  \caption{The architecture of our EfficientMotionPro network.}
  \label{fig:motionpronetwork}
\end{figure}

\subsection{Network Architecture}  
Fig.~\ref{fig:motionpronetwork} illustrates the architecture of our EfficientMotionPro network. Each frame produces an input tensor of size $[B,4,N]$ constructed from $N$ keypoints: two channels encode distance 
(or distance-derived) features and two channels encode residual motion. Distance and motion branches use Ghost~\cite{r55} 
modules for lightweight embeddings and depthwise separable convolutions; motion features are refined with ECA~\cite{r56} 
attention. Features are fused via a Lightweight Fusion Block, aggregated with attention weights, and decoded by a two-layer 
MLP to yield residuals $[B,2]$. The final displacements are reshaped into a grid field $[H/P,\,W/P,\,2]$ and smoothed by 
median pooling. Table~\ref{tab:A1} details the configuration.

\begin{table*}[t]
\centering
\caption{Architecture of the proposed \textbf{EfficientMotionPro} module. 
The input tensor $[B,4,N]$ is split into distance and motion channels, embedded via Ghost modules, enhanced with ECA, fused, 
and finally decoded into grid displacements.}
\begin{tabular}{l c c}
\hline
\textbf{Component} & \textbf{Operation} & \textbf{Output Size} \\
\hline
Input & Distance \& Motion channels & $[B,4,N]$ \\
Distance branch & GhostModule + DWConv + PWConv & $[B,48,N]$ \\
 & Feature reduction & $[B,24,N]$ \\
 & Distance attention extractor & $[B,1,N]$ \\
Motion branch & GhostModule + DWConv + PWConv & $[B,48,N]$ \\
 & ECA attention enhancement & $[B,48,N]$ \\
Fusion & Concat (Dist+Motion) $\rightarrow$ FusionBlock & $[B,48,N]$ \\
Aggregation & Attention-weighted sum & $[B,48]$ \\
Decoder & 2-layer MLP & $[B,2]$ \\
Reshape + MedianPool & Grid displacement field & $[H/P,\,W/P,\,2]$ \\
\hline
\end{tabular}
\label{tab:A1}
\end{table*}

\begin{figure}[t]
  \centering
  \includegraphics[width=0.45\textwidth]{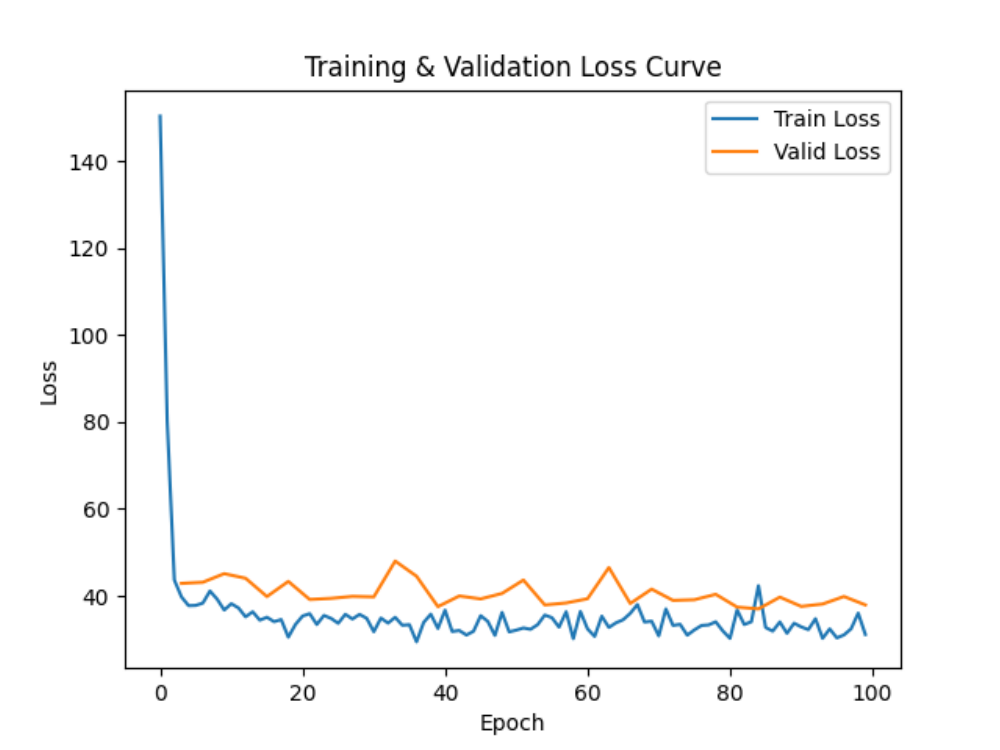}
  \caption{Training curves of \textbf{EfficientMotionPro}: the blue line shows training loss and the orange line shows validation loss.}
  \label{fig:motionpro_loss}
\end{figure}

\paragraph{(1) Implementation Details}  
We train \textbf{EfficientMotionPro} on an in-house unstable video dataset~\cite{r59,r60,r65} with train/val splits. 
Batch size is $64$; optimizer is Adam $(\beta_1=0.9,\beta_2=0.999)$ with weight decay $1\!\times\!10^{-5}$. 
Learning rate follows OneCycleLR, peaking at $2\!\times\!10^{-4}$ with $30\%$ warm-up, then cosine decay. 
Training is conducted for $100$ epochs on a single NVIDIA RTX 4090 GPU, taking approximately $12$ hours. 
We save checkpoints every $3$ epochs and keep the best validation loss as shown in Fig.~\ref{fig:motionpro_loss}.

\paragraph{(2) Complexity Analysis}  

\paragraph{Parameter Count}  
Let $C_e$ be the embedding dimension (default $C_e=48$). 
Core components include Ghost modules, depthwise separable convolutions (DW), pointwise convolutions (PW, kernel $1$), and 
ECA attention. At $C_e=48$, the whole module has $\sim 2.3 \times 10^4$ ($\approx 22.9$K) parameters, significantly fewer 
than those in typical full-frame 2D CNN or Transformer-based propagation heads.

\paragraph{Time Complexity}  
Let $N$ be the number of keypoints per frame. 
Since convolutions, softmax, and weighted aggregation are along the keypoint dimension with small kernels, the dominant MACs 
come from PW convolutions:
\begin{equation}
	\label{eq:motionpro_complexity}
	\begin{split}
	&\mathcal{O}(N C_e^2) \;[\text{PW: dist reduction } C_e \!\to\! C_e/2] \\
	&+ \mathcal{O}(4 N C_e^2) \;[\text{PW: motion } 2C_e \!\to\! 2C_e] \\
	&+ \mathcal{O}(N C_e^2) \;[\text{PW: motion } C_e \!\to\! C_e] \\
	&+ \mathcal{O}(N C_e k) \;[\text{DW blocks}] 
	\;+\; \mathcal{O}(N) \;[\text{aggregation}] \\
	&\approx \mathcal{O}(5 N C_e^2).
	\end{split}
\end{equation}
Thus the cost scales linearly with $N$ and is largely decoupled from image resolution or grid size; 
the cost of median pooling is negligible due to its small kernel size.

\paragraph{Memory Complexity}  
Intermediate tensors scale as $\mathcal{O}(B C_e N)$ (independent of $HW$).

\paragraph{Complexity Estimation}  
Numerical MAC estimates (ignoring BN/activations/ECA) with $C_e=48$ are shown in Table~\ref{tab:A2}.
\begin{table}[t]
\centering
\caption{Estimated MACs per frame of EfficientMotionPro under different keypoint counts $N$.}
\begin{tabular}{c c}
\hline
\textbf{Keypoints $N$} & \textbf{Estimated MACs / frame} \\
\hline
$128$ & $\approx 2.14$M \\
$256$ & $\approx 4.28$M \\
$512$ & $\approx 8.55$M \\
\hline
\end{tabular}
\label{tab:A2}
\end{table}

\section{OnlineSmoother Details}

\subsection{Training Objectives}  
In dynamic scenes or under low-frame-rate conditions, grid vertex trajectories often exhibit local oscillations due 
to jitter, occlusion, or non-rigid disturbances, which compromise temporal consistency and introduce geometric 
distortions. To achieve online stabilization with minimal latency and without using future frames, we introduce 
the \textbf{OnlineSmoother} network. It leverages only the historical trajectory data from a sliding buffer, 
employing a Lite LS-3D~\cite{r57} encoder to extract spatio-temporal features and a Star-gated~\cite{r58} decoder 
to predict dynamic smoothing kernels for the $x$ and $y$ directions separately. This kernel-based iterative update 
mechanism progressively suppresses high-frequency oscillations while preserving motion intentionality. 
We additionally employ spatial consistency and projection consistency objectives for training.

\paragraph{(1) Spatial Distortion Constraint}  
To maintain the local geometric structure of the mesh during temporal transformation, we introduce a distortion loss 
based on triangular mesh elements. The smoothed trajectories are mapped back to the grid coordinate system to 
construct triangular structures (eight varying configurations per quad). The loss penalizes deviations in edge 
length ratios and angles from their initial reference state:
\begin{equation}
	\label{eq:spatial_loss}
	\begin{split}
	\mathcal{L}_{\text{spatial}} 
	&= \tfrac{1}{|G|} \sum_{i,j} \Bigg[ 
	\lambda_{\text{edge}} \sum_{\text{edges } e} 
	\sqrt{ \left( \frac{\lVert e \rVert}{\lVert e_0 \rVert} - 1 \right)^2 + \epsilon^2 } \\
	&\quad + \lambda_{\text{angle}} \sum_{\text{angles } \theta} 
	\sqrt{ \left( \frac{\theta}{\theta_0} - 1 \right)^2 + \epsilon^2 } \Bigg]
	\end{split}
\end{equation}
where $e_0$ and $\theta_0$ represent the edge lengths and angles of the triangles in the initial, undeformed 
reference mesh $g_{\text{ref}}$, and $|G|$ is the total number of grid cells.

\paragraph{(2) Keypoint Projection Consistency Constraint}  
To further ensure the alignment between the smoothed motion field and the observed keypoint trajectories, we 
introduce a homography-based projection supervision. Specifically, local homography transformations are constructed 
from the smoothed grid motion field $S_t$. The original keypoint positions are then projected using these 
homographies, and a consistency error is computed against their actual warped positions:
\begin{equation}
	\label{eq:project_loss}
	\mathcal{L}_{\text{proj}} = \tfrac{1}{|\Omega_{kp}|} \sum_{p \in \Omega_{kp}} \omega_{t,p} \, 
	\rho\!\left( \mathcal{W}(p; O_t),\, H_{t}^{\text{local}}(p; S_t) \cdot p \right),
\end{equation}
where $\rho(\cdot, \cdot)$ denotes the Charbonnier penalty $\rho(a, b) = \sqrt{\lVert a - b \rVert_2^2 + \epsilon^2}$,
and $H_{t}^{\text{local}}(p; S_t)$ represents the local homography estimated from the smoothed grid $S_t$ at point $p$.
This term effectively enhances the accuracy and geometric plausibility of the smoothed trajectory field at the 
locations of real keypoints.

\paragraph{(3) Total Loss}  
The final optimization objective for the trajectory smoothing module is defined as the weighted combination of all 
\textbf{three} losses:
\begin{equation}
	\label{eq:smooth_total_loss}
	\mathcal{L}_{\text{total}} = \mathcal{L}_{\text{temp}} 
	+ \lambda_{\text{spatial}}\mathcal{L}_{\text{spatial}} 
	+ \lambda_{\text{proj}}\mathcal{L}_{\text{proj}},
\end{equation}
where $\lambda_{\text{time}}=20$, $\lambda_{\text{freq}}=1$, $\lambda_{\text{spatial}}=10$, and 
$\lambda_{\text{proj}}=5$ are the default hyperparameters inside/alongside $\mathcal{L}_{\text{temp}}$ (time/freq) and the two auxiliary terms.

\begin{figure}[t]
  \centering
  \includegraphics[width=0.45\textwidth]{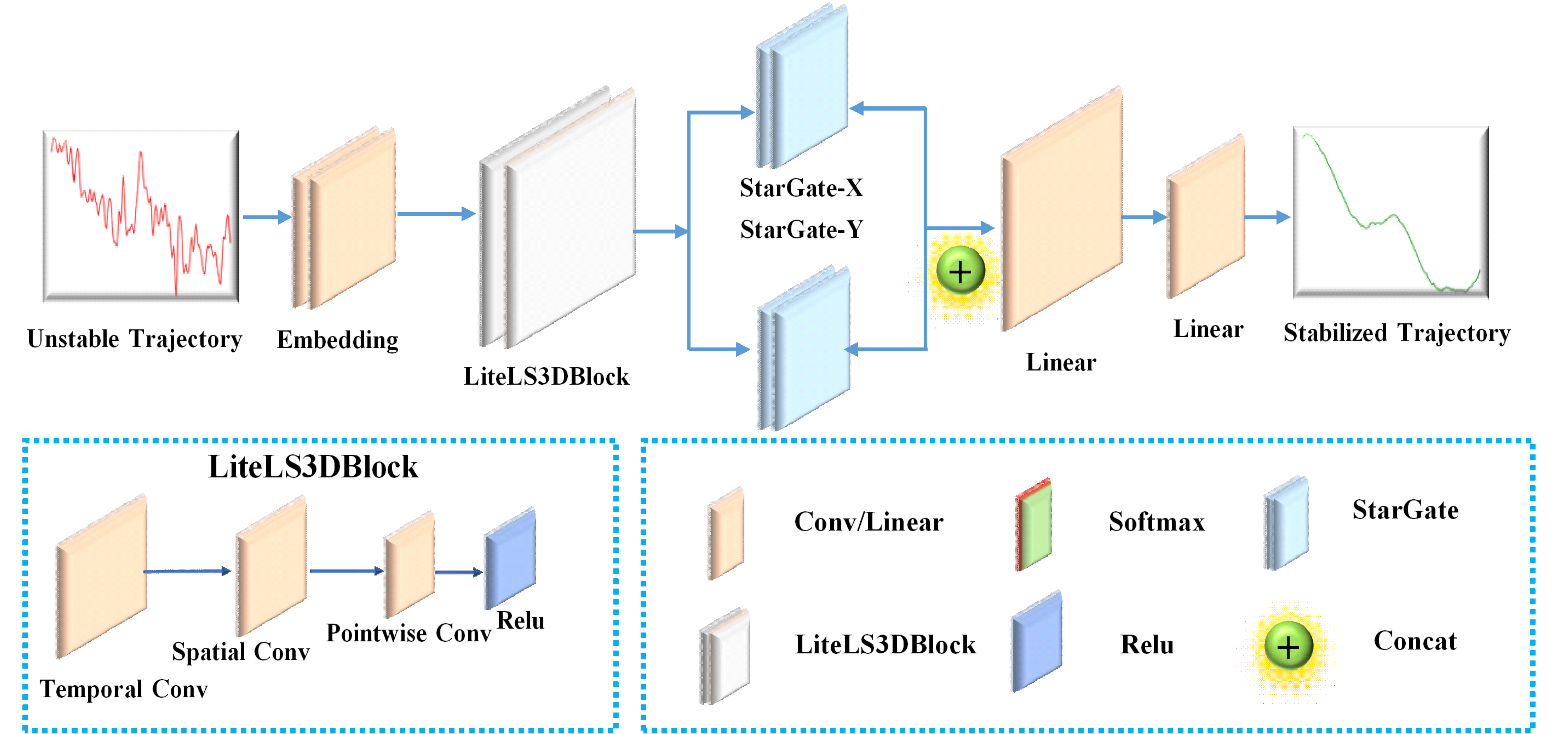} 
  \caption{The architecture of our OnlineSmoother network.}
  \label{fig:onlinesmoothnetwork}
\end{figure}

\subsection{Network Architecture}  
Fig.~\ref{fig:onlinesmoothnetwork} illustrates the architecture of our OnlineSmoother network. Given frame-wise 
displacements, we form a causal trajectory tensor $[B,2,T,H,W]$ (two channels for $x/y$). \textbf{OnlineSmoother} 
consists of: trajectory encoder (Linear $(2\!\to\!64)$ + ReLU), spatiotemporal modeling (Lite LS-3D~\cite{r57}: 
temporal large-kernel DW-Conv3d + spatial $1\!\times\!3\!\times\!3$ DW-Conv3d, optional 
$1\!\times\!1\!\times\!1$ PW-Conv3d for mixing), Star-gated~\cite{r58} dynamic-kernel decoder (two linear heads 
for gating/amplitude and one scale head), and kernel-guided iterative smoothing. Table~\ref{tab:smoother_arch} 
summarizes the design.

\begin{table*}[t]
\centering
\caption{Architecture of the \textbf{OnlineSmoother} module. Per-vertex processing; outputs are per-vertex kernels 
and smoothed trajectories.}
\begin{tabular}{l c l}
\hline
\textbf{Submodule} & \textbf{Tensor Shape (Input $\to$ Output)} & \textbf{Structural Components} \\
\hline
Trajectory Encoder & $[B,2,T,H,W] \to [B,64,T,H,W]$ & Linear $(2\!\to\!64)$ + ReLU \\
Lite LS-3D (Temporal) & $[B,64,T,H,W] \to [B,64,T,H,W]$ & DW-Conv3d $(k_t\!\times\!1\!\times\!1$, dilation $d_t)$ + ReLU \\
Lite LS-3D (Spatial) & $[B,64,T,H,W] \to [B,64,T,H,W]$ & DW-Conv3d $(1\!\times\!3\!\times\!3)$ + ReLU \\
(Optional) Channel Mixing & $[B,64,T,H,W] \to [B,64,T,H,W]$ & PW-Conv3d $(1\!\times\!1\!\times\!1$, groups=4) + ReLU \\
Kernel Decoder (x) & $[64]\to[6]$ (per vertex) & Linear (gate), Sigmoid; Linear (amplitude) \\
Kernel Decoder (y) & $[64]\to[6]$ (per vertex) & Same as $x$ \\
Scale Decoder (shared) & $[64]\to[1]$ (per vertex) & Linear (scale) \\
Kernel Map & $[B,12,T,H,W]$ & Element-wise mult., Star-gated fusion \\
Kernel-guided Smoothing & $[B,2,T,H,W]\to[B,2,T,H,W]$ & Iterative consistency refinement (6-step kernel) \\
\hline
\end{tabular}
\label{tab:smoother_arch}
\end{table*}

\paragraph{(1) Implementation Details}  
We train and evaluate \textbf{OnlineSmoother} on the same self-collected unstable video dataset (with sequence-level 
train/val splits) that is also used for \textbf{EfficientMotionPro}.

Batch size is $1$ to preserve causality. 
Optimizer: Adam $(\beta_1=0.9,\beta_2=0.99)$, weight decay $1\times10^{-5}$. 
Learning rate: OneCycleLR with maximum $2\times10^{-4}$, warm-up for the first $30\%$ steps, then cosine annealing. 
The smoothing loss combines $\mathcal{L}_{\text{time}}$, $\mathcal{L}_{\text{spatial}}$, and $\mathcal{L}_{\text{proj}}$ with default 
weights $(\lambda_{\text{time}},\lambda_{\text{spatial}},\lambda_{\text{proj}})=(20,10,5)$ and internal $\lambda_{\text{freq}}=1$. 
Training runs for $100$ epochs; gradient clipping (threshold $5.0$) ensures numerical stability. 

\begin{figure}[t]
  \centering
  \includegraphics[width=0.45\textwidth]{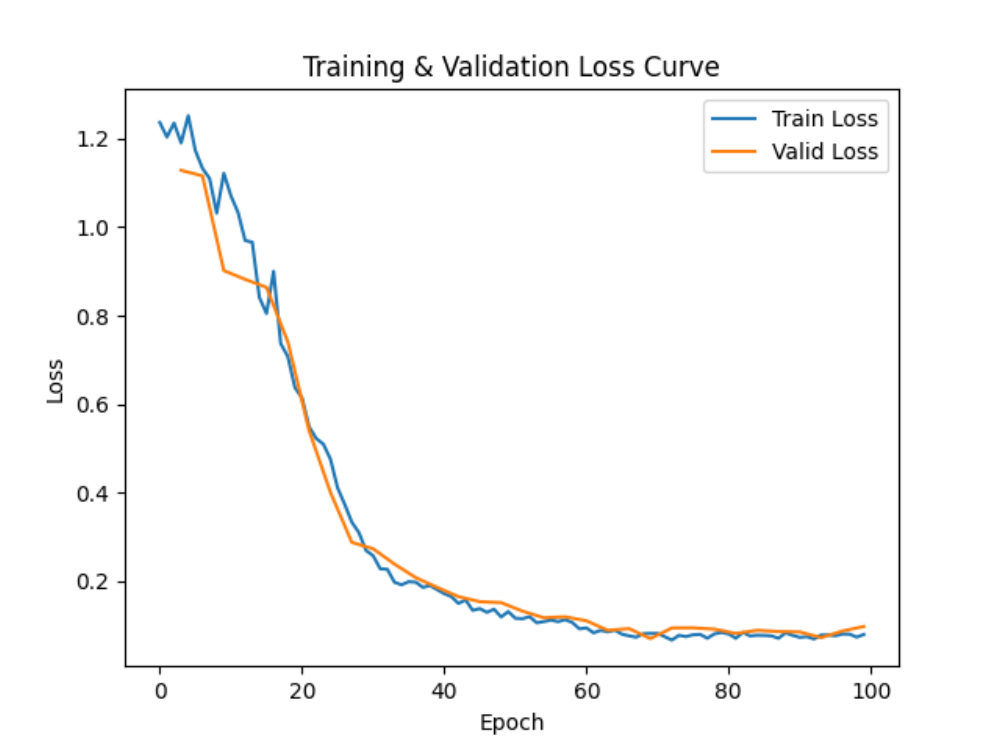}
  \caption{Training curves of \textbf{OnlineSmoother}: blue for training loss, orange for validation loss.}
  \label{fig:onlinesmooth_loss}
\end{figure}

Validation is performed every $3$ epochs; the best model based on the validation loss (Fig.~\ref{fig:onlinesmooth_loss}) is saved.  
Training is conducted on a single NVIDIA RTX 4090 GPU (about $2.5$ hours).

\paragraph{(2) Complexity Analysis}  

\paragraph{Parameter Count.}  
Let $C=64$ (embedding dim), temporal kernel $k_t=5$, and optional PW mixing controlled by \texttt{use\_pointwise}. 
Counts (approx.): Encoder Linear $(2\!\to\!C)$: $2C$; 
DW-temporal: $\mathcal{O}(Ck_t)$; 
DW-spatial: $\mathcal{O}(9C)$; 
(Opt.) PW-Conv3d (groups=4): $\approx C^2/4$; 
Star-gated decoder: two Linear $(C\!\to\!6)$ per direction ($2\cdot6C$ each) + one Linear $(C\!\to\!1)$ scale. 
Total (no PW): for $C=64,k_t=5$ is $\sim2.6$K; with PW (groups=4) adds $\sim1.0$K, totaling $\sim3.6$K.

\paragraph{Time Complexity.}  
Let batch $B$, temporal window $T$, grid size $H\times W$.
Dominant MACs (ignoring BN/activations/Sigmoid) come from per-vertex Linear and depthwise 3D convolutions:
\begin{equation}
	\label{eq:time_complexity}
	\mathcal{O}\!\big(BTHW \cdot (2C + Ck_t + 9C)\big)
\end{equation}
plus decoder heads $\approx\mathcal{O}(BTHW\cdot 25C)$ and (opt.) PW mixing $\mathcal{O}(BTHW\cdot C^2/\text{groups})$. 
Overall (no PW):
\begin{equation}
	\label{eq:smoothnet_complexity}
	\mathcal{O}\!\big(BTHW\cdot (2C + Ck_t + 9C + 25C)\big).
\end{equation}
Kernel-guided updates with $R$ rounds and six history steps are scalar-weighted and negligible relative to the forward pass (even with $R=20$).

\paragraph{Memory Complexity.}  
Feature maps are $[B,C,T,H,W]$; kernel heads $[B,6,T,H,W]$ and $[B,1,T,H,W]$. 
Thus memory is $\mathcal{O}(BCTHW)$, independent of input image resolution.

\paragraph{Numerical Estimation (Default $C=64, k_t=5, T=7$).}  
Ignoring BN/activations/Sigmoid and with $B=1$:
\begin{equation}
	\label{eq:macs_estimation}
	\mathrm{MACs} \approx T \cdot H \cdot W \cdot C \cdot (k_t + 35)
\end{equation}
For $H=W=32$:
\begin{equation}
	\label{eq:macs_example}
	7 \cdot 32 \cdot 32 \cdot 64 \cdot 40 \;\approx\; 18.4\,\mathrm{M}
\end{equation}
We summarize per-grid costs in Table~\ref{tab:OnlineSmoother_complexity}.
\begin{table}[t]
\centering
\caption{Estimated MACs of the OnlineSmoother backbone at different grid sizes ($C=64$, $k_t=5$, $T=7$, $B=1$).}
\begin{tabular}{c c c}
\hline
\textbf{Window $T$} & \textbf{Grid $H \times W$} & \textbf{MACs / frame} \\
\hline
$7$ & $16 \times 16$ & $\approx 4.6$M \\
$7$ & $32 \times 32$ & $\approx 18.4$M \\
$7$ & $48 \times 48$ & $\approx 41.5$M \\
\hline
\end{tabular}
\label{tab:OnlineSmoother_complexity}
\end{table}


\section{Multi-Threaded Asynchronous Buffering Pipeline: Additional Details}
\label{supp:multithreading}

\paragraph{(1) Motivation and Formulation}  
Although each individual module is computationally lightweight, executing them in sequence leads to the accumulation 
of stage latencies:
\begin{equation}
	\label{eq:serial_time}
	T_{\text{serial}} = t_{\text{est}} + t_{\text{prop}} + t_{\text{smooth}},
\end{equation}
where \(t_{\text{est}}\), \(t_{\text{prop}}\), and \(t_{\text{smooth}}\) denote the average per-frame runtimes of 
the \textbf{Motion Estimation}, \textbf{Motion Propagation}, and \textbf{Motion Compensation} stages, respectively. 
This serial bottleneck prevents strict real-time performance.

\paragraph{(2) Pipeline Architecture}  
To mitigate this, we decompose the stabilization pipeline into three concurrent daemon threads, decoupled by bounded 
FIFO (First-In-First-Out) shared queues with back-pressure (threads block on empty/full states). This design ensures 
safe resource handling even under exceptional conditions:
\begin{itemize}
    \item \textbf{T\textsubscript{ME} (Motion Estimation):} Acquires input frames \(I_t\), performs keypoint 
	collaboration detection \(\mathcal{D}\), homogenization via SSC, and sparse keypoint-guided causal flow fusion 
	to produce \(\mathbf{m}_t = [x_{kp}; y_{kp}; u; v]\) and the reweighted flow field 
	\(\hat{\mathbf{f}}_{t \leftarrow t-1}\).
    \item \textbf{T\textsubscript{MP} (Motion Propagation):} Consumes outputs from T\textsubscript{ME}. 
	Executes the EfficientMotionPro network, which computes the multi-homography prior, predicts per-vertex 
	residuals \(\Delta g_{\text{res},t}\) via the lightweight backbone, and outputs the full-frame grid motion field 
	\(\Delta g_t\).
    \item \textbf{T\textsubscript{MC} (Motion Compensation):} Consumes grid motions \(\Delta g_t\) (or integrated 
	trajectories \(O_t\)) from T\textsubscript{MP}. Applies the \textbf{OnlineSmoother} network for online causal smoothing, 
	yielding the stabilized trajectory \(S_t\). Finally, it computes the compensation field \(M_t = S_t - O_t\), 
	performs frame warping via \(\mathcal{W}(x; M_t)\), and outputs the stabilized frame \(\tilde I_t\).
\end{itemize}

Threads communicate exclusively through the following shared queues:
\begin{equation}
	\label{eq:queue_definition}
	\renewcommand{\arraystretch}{1.2}
	\begin{array}{l @{\quad} l}
	\multicolumn{2}{l}{Q_{\text{ME}\to\text{MP}} \colon} \\
	& \text{Transmits } \mathbf{m}_t \text{ (keypoint motion features)}, \\
	& \text{plus metadata}, \\
	\multicolumn{2}{l}{Q_{\text{MP}\to\text{MC}} \colon} \\
	& \text{Transmits } \Delta g_t \text{ (grid motion field)}, \\
	& \text{plus metadata}.
	\end{array}
\end{equation}

\paragraph{(3) Theoretical Analysis}  
Let the steady-state per-frame service times be \(\mathbf{t}=(t_{\text{est}}, t_{\text{prop}}, t_{\text{smooth}})\) 
and the capacity of each queue be \(C\):
\begin{itemize}
    \item \textbf{Throughput (Steady-State):} The pipeline's frame rate is bounded by the slowest stage.
	\begin{equation}
		\label{eq:pipeline_fps}
		\begin{aligned}
		\eta_{\text{pipe}} &\approx \frac{1}{\max\{t_{\text{est}}, t_{\text{prop}}, t_{\text{smooth}}\}},\\
		\text{FPS}_{\text{max}} &\approx \frac{1}{\max\{t_{\text{est}}, t_{\text{prop}}, t_{\text{smooth}}\}}.
		\end{aligned}
	\end{equation}

    \item \textbf{Theoretical Speedup (vs. Serial Execution):}
	\begin{equation}
		\label{eq:speedup}
		S = \frac{t_{\text{est}} + t_{\text{prop}} + t_{\text{smooth}}}
				{\max\{t_{\text{est}}, t_{\text{prop}}, t_{\text{smooth}}\}}.
	\end{equation}

    \item \textbf{End-to-End Latency:} The delay for a frame to traverse the entire pipeline is reduced from 
	the serial sum to the duration of the longest stage plus queuing delays, which are minimal for adequately 
	sized buffers.
    \item \textbf{Memory Overhead:} The memory footprint is dominated by the bounded queues.
	\begin{equation}
		\label{eq:memory_overhead}
		\mathcal{M} = \mathcal{O}\!\big(
		C_{\text{ME}} \cdot |\mathbf{m}_t|
		+ C_{\text{MP}} \cdot |\Delta g_t|
		\big),
	\end{equation}
    where \(|\mathbf{m}_t|\) and \(|\Delta g_t|\) represent the size of the data packets transmitted between 
	threads, and \(C_{\text{ME}}, C_{\text{MP}}\) are the capacities of the respective queues.
\end{itemize}

\paragraph{(4) Complexity and Real-Time Properties}  
\begin{itemize}
    \item \textbf{Computational Complexity:} Pipeline parallelism does not change the asymptotic complexity 
	\(\mathcal{O}(\cdot)\) of each individual stage; it primarily reduces the wall-clock time from the sum 
	\(t_{\text{est}} + t_{\text{prop}} + t_{\text{smooth}}\) to the maximum \(\max\{t_{\text{est}}, t_{\text{prop}}, 
	t_{\text{smooth}}\}\).
    \item \textbf{Memory/Bandwidth:} The memory and inter-thread communication bandwidth scale linearly with the 
	chosen queue capacities. This overhead is typically negligible compared to the memory required for processing 
	batches of frames or implementing global buffering and reordering strategies.
    \item \textbf{Real-Time Performance:} If the service times are balanced (\(t_{\text{est}} \approx 
	t_{\text{prop}} \approx t_{\text{smooth}}\)), the throughput approaches a \(3\times\) speedup over serial 
	execution. If one stage is the dominant bottleneck, the overall throughput is determined by that stage's 
	processing rate, while the latency for an individual frame is still reduced to approximately the duration 
	of that longest stage.
\end{itemize}

\section{Frame Outpainting Details}

When the stabilized video is cropped, we first compute the cropping ratio and then determine the scaling required for outpainting. The scaling factors are derived from the cropped dimensions defined by the frame borders.

\paragraph{Cropping Ratio and Cropped Size}
Let \(W_{\text{orig}}, H_{\text{orig}}\) be the original width and height. Let \(B_{\text{hor}}\) and \(B_{\text{ver}}\) denote the per-side horizontal (left/right) and vertical (top/bottom) border thickness (in pixels), respectively. Then
\begin{equation}
\label{eq:border_crop}
\begin{aligned}
W_{\text{cropped}} &= W_{\text{orig}} - 2\,B_{\text{hor}},\\
H_{\text{cropped}} &= H_{\text{orig}} - 2\,B_{\text{ver}}.
\end{aligned}
\end{equation}
The cropping ratio (content preservation) is
\begin{equation}
\label{eq:crop_ratio}
C \;=\; \frac{W_{\text{cropped}}\,H_{\text{cropped}}}{W_{\text{orig}}\,H_{\text{orig}}}\in(0,1].
\end{equation}

\paragraph{Scaling Factors for Outpainting}
To compensate for the cropped regions, we define the anisotropic scaling factors
\begin{equation}
\label{eq:scale_w}
\text{scale}_w \;=\; \frac{W_{\text{orig}}}{W_{\text{cropped}}}
= \frac{W_{\text{orig}}}{W_{\text{orig}} - 2\,B_{\text{hor}}},
\end{equation}
\begin{equation}
\label{eq:scale_h}
\text{scale}_h \;=\; \frac{H_{\text{orig}}}{H_{\text{cropped}}}
= \frac{H_{\text{orig}}}{H_{\text{orig}} - 2\,B_{\text{ver}}}.
\end{equation}
If aspect ratio must be preserved (recommended), use an isotropic factor
\begin{equation}
\label{eq:scale_iso}
s \;=\; \max\!\big(\text{scale}_w,\;\text{scale}_h\big)\;\;\;(\,s\ge 1\,).
\end{equation}

\begin{figure}[t]
  \centering
  \includegraphics[width=0.45\textwidth]{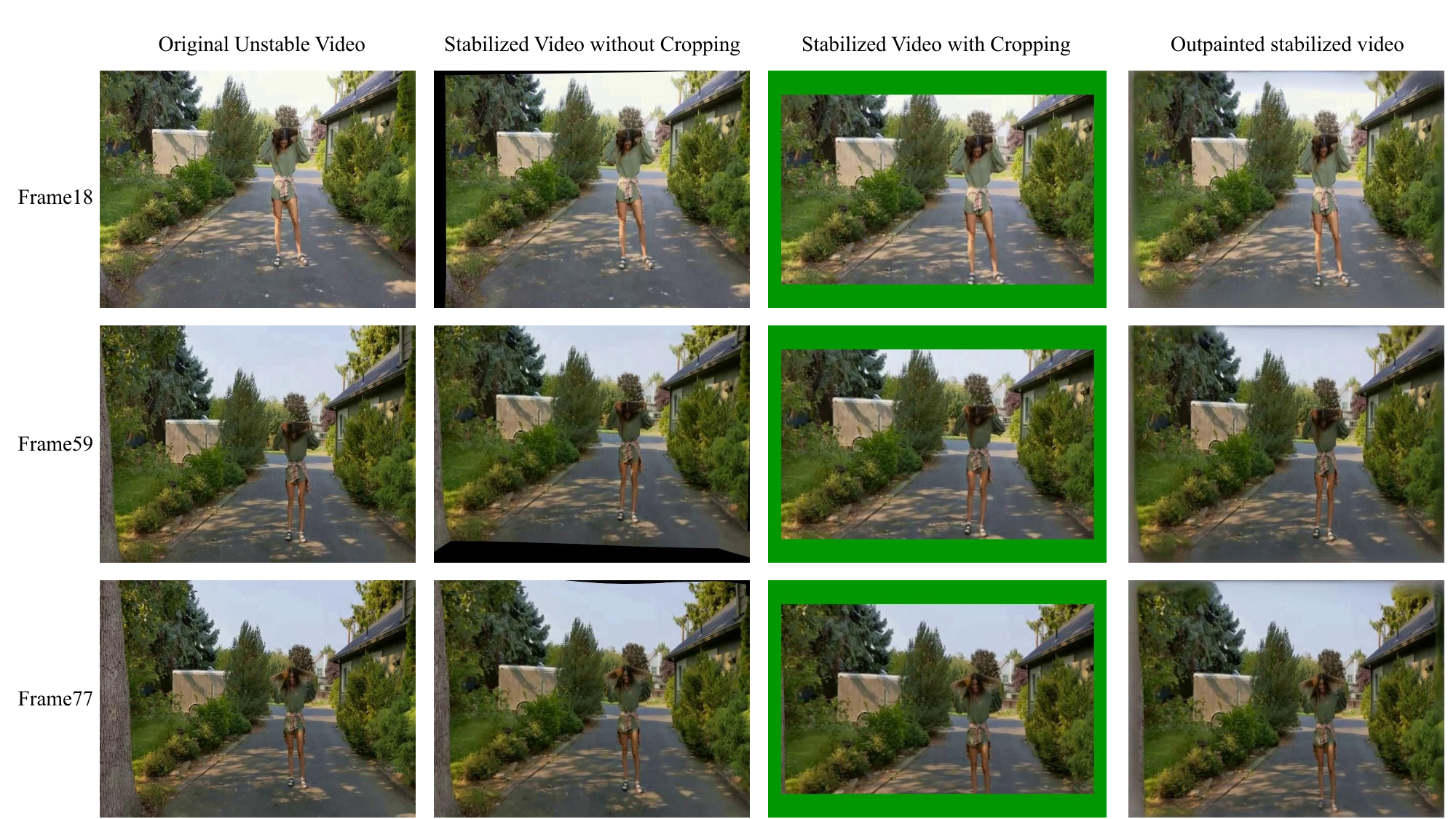}
  \caption{Left to right: Original unstable video; Stabilized (uncropped); Stabilized (cropped); Stabilized (with frame outpainting).}
  \label{fig:vieo_outpainting}
\end{figure}

\paragraph{Outpainting with the ProPainter Model}
ProPainter~\cite{r71} is then applied to the missing boundary regions induced by the target scaling. In practice, we first warp each stabilized frame to the outpainted canvas using the chosen factor (\(\text{scale}_w,\text{scale}_h\) or \(s\)), and then invoke ProPainter to fill the uncovered areas near the borders. This preserves the original content while minimizing visible seams at frame boundaries. Figures~\ref{fig:vieo_outpainting} illustrates the difference between the pre- and post-stabilization video with outpainting.

\begin{figure*}[t]
  \centering
  \includegraphics[width=0.9\textwidth]{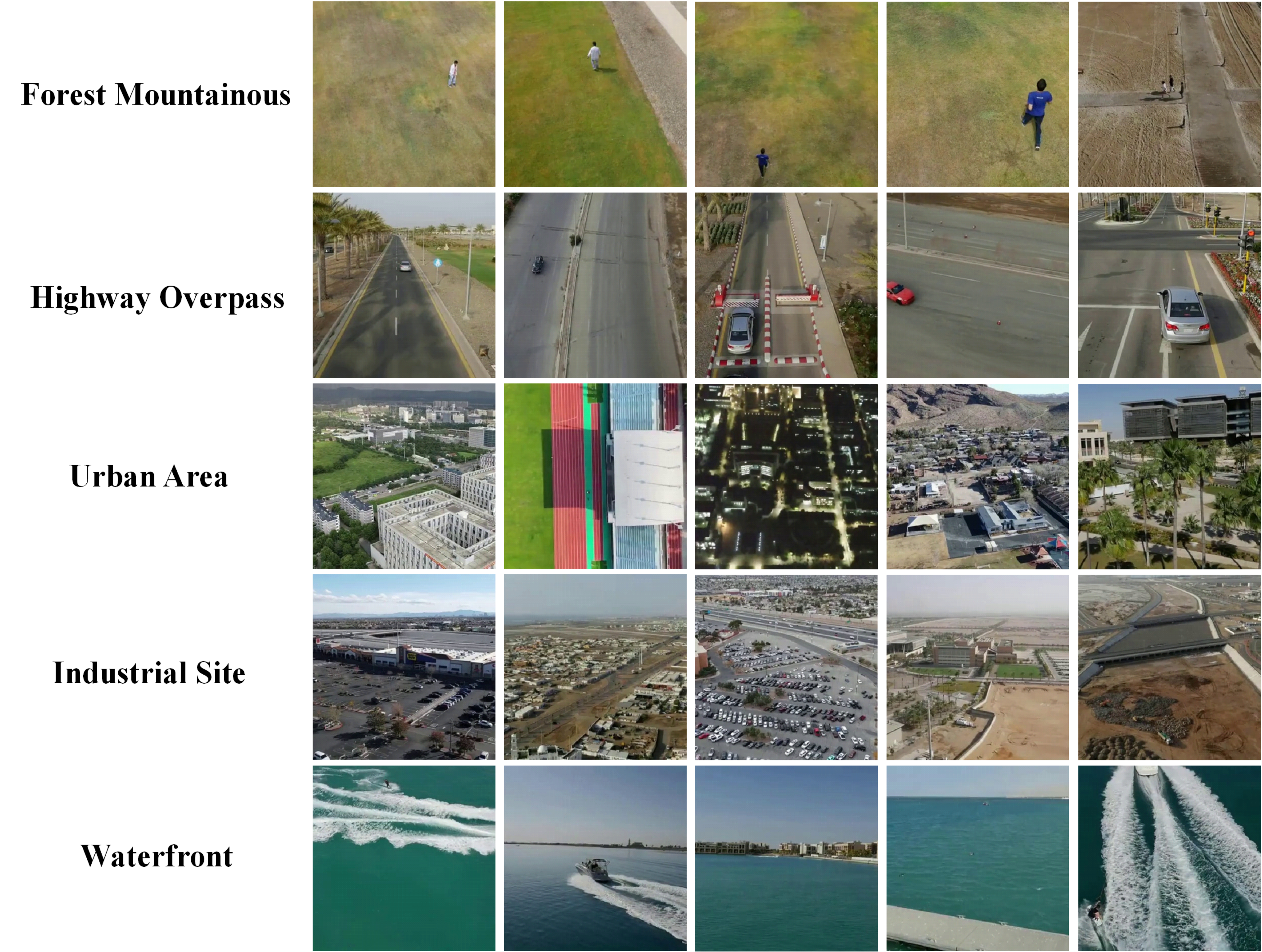}
  \caption{Representative examples of unstable UAV visible-light videos from diverse scenarios, including urban 
  areas, highways, forested mountains, waterfronts, and industrial sites.}
  \label{fig:UAV_test_visible}
\end{figure*}

\begin{figure*}[t]
  \centering
  \includegraphics[width=0.9\textwidth]{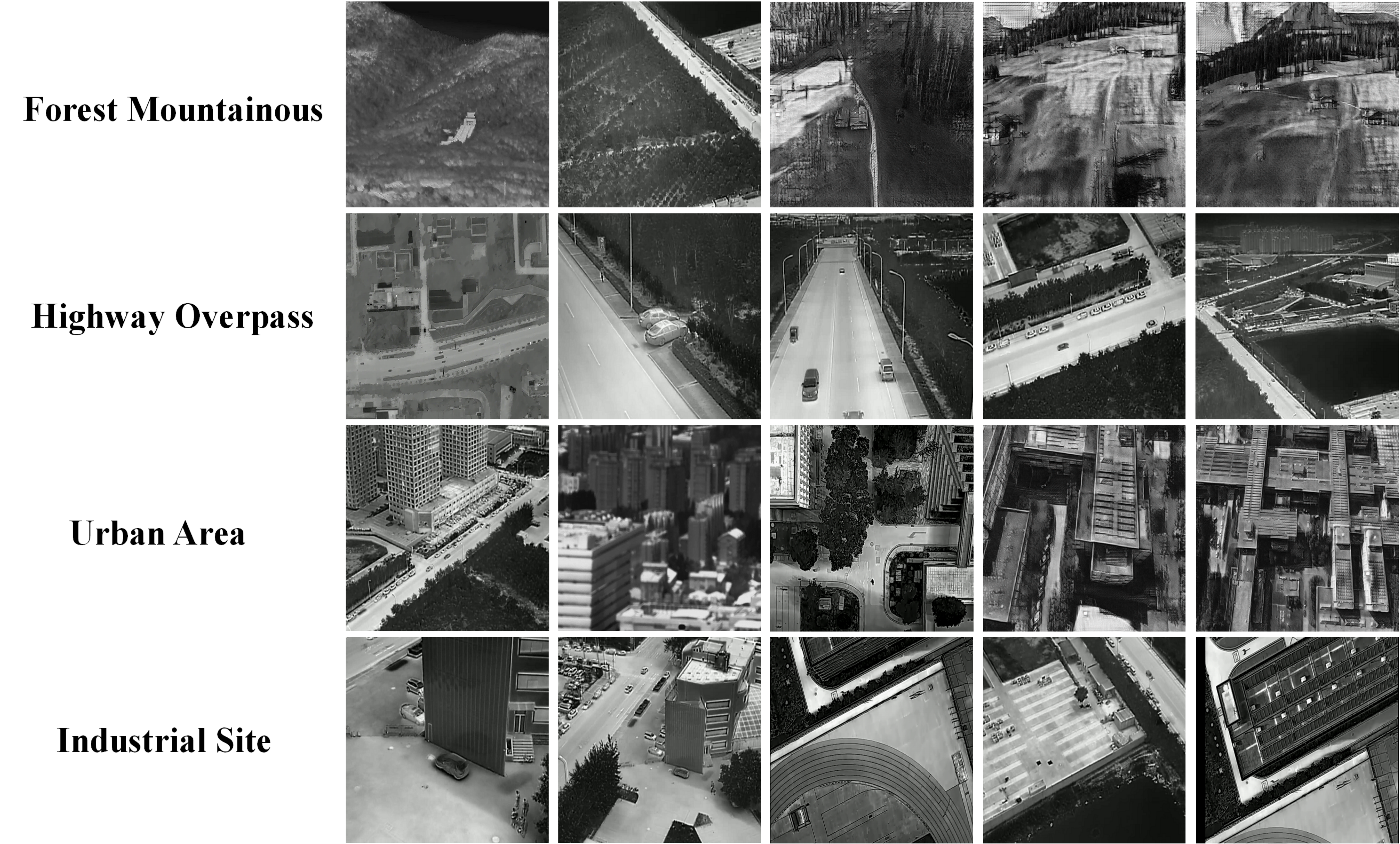}
  \caption{Representative examples of unstable UAV infrared videos from diverse scenarios, including urban areas, 
  highways, forested mountains, and industrial sites.}
  \label{fig:UAV_test_infrared}
\end{figure*}

\section{Additional Details for UAV-Test Dataset}
To comprehensively evaluate robustness and generalization in complex real-world conditions, we construct and release 
the \textbf{UAV-Test} dataset for video stabilization. This dataset captures instability patterns common in aerial 
operations, such as high-frequency vibrations, aggressive maneuvers, and occlusions. 
Figures~\ref{fig:UAV_test_visible} and \ref{fig:UAV_test_infrared} present sample visible-light and infrared scenes.

\paragraph{Data Collection Protocol}
Data were captured using a DJI Matrice~300 RTK UAV equipped with a DJI Zenmuse H20N gimbal camera, featuring both 
a wide-angle visible-light sensor and a long-wave infrared sensor. The visible-light lens supports optical zoom from 
4.5--360~mm, while the infrared lens (8--14~$\mu$m) has a fixed focal length of 150~mm. The shutter speed was set to 
automatic, and ISO was set within 100--1600. Videos were recorded at 1280$\times$720 and 1920$\times$1080 resolutions, 
with frame rates of 25--30~FPS, simulating varying transmission bandwidths and computational loads.

\paragraph{Scene Distribution}
UAV-Test comprises 92 video sequences totaling $\sim$26 minutes, covering five categories:
\begin{itemize}
    \item \textbf{Urban Areas} (19 sequences, 4.5~min): Building circumnavigation and traversal under wind levels 
	3--4.
    \item \textbf{Highways} (22 sequences, 6.3~min): High-speed parallel tracking under wind levels 2--3.
    \item \textbf{Forested Mountains} (18 sequences, 5.9~min): Canopy penetration and terrain tracking under wind 
	levels 3--5.
    \item \textbf{Waterfronts} (20 sequences, 6.0~min): Disturbance from water-surface reflections under wind 
	levels 3--4.
    \item \textbf{Industrial Sites} (13 sequences, 3.2~min): Occlusions from machinery under wind level~3.
\end{itemize}

\paragraph{Annotation and Cleaning}
Corrupted frames caused by transmission errors were automatically detected and removed using \texttt{ffmpeg}, 
ensuring clean sequences for evaluation.

\paragraph{Partitioning and Licensing}
UAV-Test is designated as a pure testing set. A stratified sampling strategy based on scene type, weather 
(sunny/rainy/foggy), and time of day (day/night) ensures diversity. The dataset will be released with source code; 
access can be requested via email. All flights complied with airspace regulations, and privacy-sensitive information 
such as human faces and license plates was blurred.

\section{Runtime Analysis}

\begin{table}[t]
\centering
\caption{Per-frame runtime comparison on Jetson AGX Orin. 
We report the average per-frame runtime (ms) and FPS. 
Our method is significantly faster than MeshFlow, StabNet, and NNDVS, 
and second only to the highly lightweight Liu et al.~\cite{r22}.}
\label{tab:runtime}
\setlength{\tabcolsep}{6pt}
\renewcommand{\arraystretch}{1.05}
\begin{tabular}{lcc}
\toprule
\textbf{Method} & \textbf{Runtime (ms)} & \textbf{FPS} \\
\midrule
MeshFlow~\cite{r23}   & 168.95 & 5.92 \\
StabNet~\cite{r10}    & 179.88 & 5.56 \\
NNDVS~\cite{r15}      & 340.02 & 2.94 \\
Liu et al.~\cite{r22} & \textcolor{red}{36.41} & \textcolor{red}{27.47} \\
Ours                  & \textcolor{blue}{78.94} & \textcolor{blue}{12.67} \\
\bottomrule
\end{tabular}
\end{table}

While many stabilization algorithms achieve real-time performance on high-end GPUs, their efficiency often drops 
sharply on embedded platforms. To assess practical efficiency, we benchmarked our method on the Jetson AGX Orin and 
compared it with MeshFlow~\cite{r23}, StabNet~\cite{r10}, NNDVS~\cite{r15}, and Liu et al.~\cite{r22}. Experimental 
results show that, thanks to its lightweight architecture and multi-threaded buffering, our method runs at $\sim$13~FPS 
(78.94~ms per frame). This is markedly faster than MeshFlow, StabNet, and NNDVS, and second only to the extremely 
lightweight approach of Liu et al., demonstrating a favorable trade-off among accuracy, real-time performance, and 
deployment adaptability. \textit{Note:} all timings exclude the optional frame-outpainting post-processing step; only the 
time to produce a stabilized frame is measured for every method.


\begin{figure}[t]
  \centering
  \includegraphics[width=0.45\textwidth]{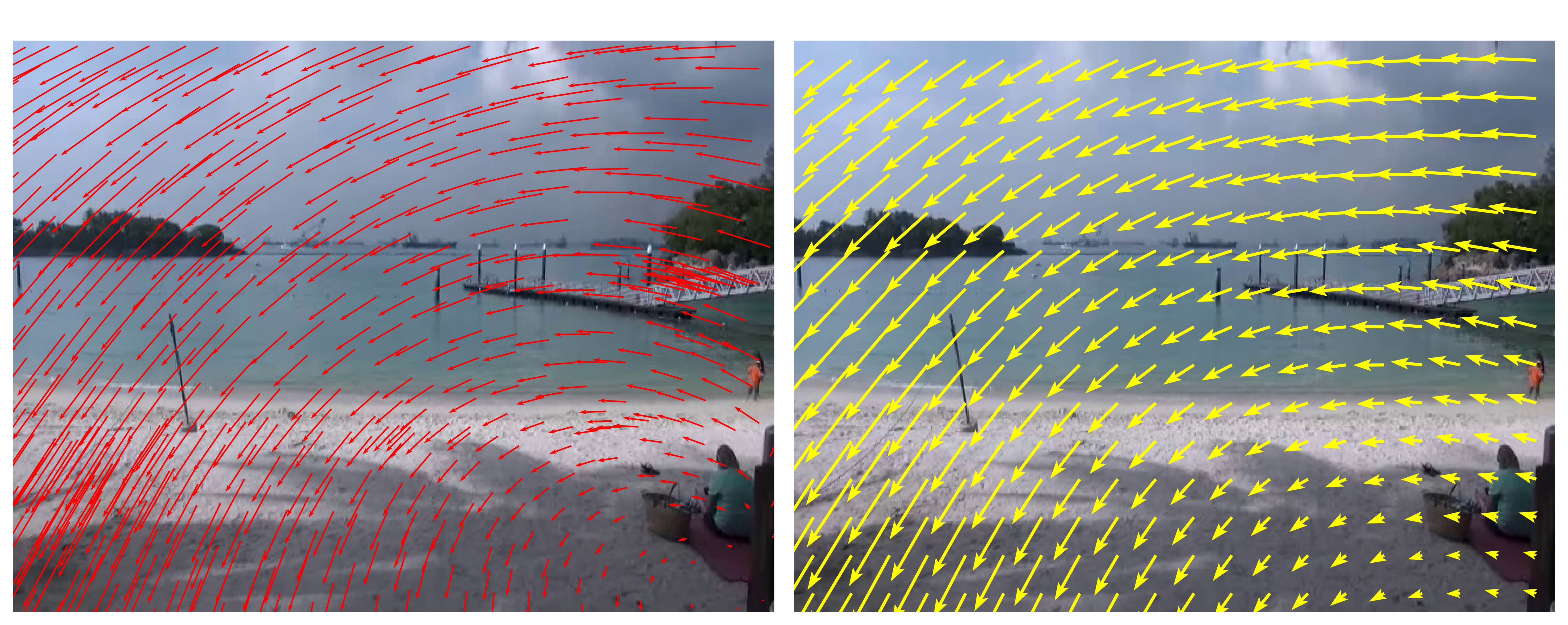}
  \caption{Visualization of motion propagation. The sparse optical flow guided by initial keypoints is 
  contrasted with the grid-wise motion after propagation by our EfficientMotionPro module. The results show 
  that our propagation yields more consistent and structured motion, robust to disturbances caused by dynamic 
  objects and noise.}
  \label{fig:visualize_motionpro}
\end{figure}

\section{Evaluation Metrics and Implementation}

For reproducibility, we explicitly define the three core metrics used in the main paper, following~\cite{r3}, 
together with implementation details.

\paragraph{Cropping Ratio (C)}  
For each frame, a global homography between the input and the stabilized output is estimated, and the scaling 
components are used to compute the cropping ratio. The overall cropping ratio \(C\) is defined as the average 
across all frames. Let the effective field of view after stabilization be \(\Omega_t\) and the original field 
of view be \(\Omega\), then:
\begin{equation}
  \label{eq:cropping}
  C = \frac{1}{T} \sum_{t=1}^T \frac{|\Omega_t|}{|\Omega|}.
\end{equation}

\noindent \textbf{Implementation details.} The cropping ratio is estimated from the scaling factors \(s_x, s_y\) of 
the frame homography \(H_t\). To obtain a stricter measure, the four image corners are projected onto the stabilized view, 
and the preserved ratios along the width and height are computed; the minimum of the two defines the FOV metric, 
capturing the worst-case preserved region.

\paragraph{Distortion Value (D)}  
The distortion value measures anisotropic scaling introduced by the affine part of the homography. Specifically, 
for each frame \(t\), let \(\sigma_{\max}(H_t)\) and \(\sigma_{\min}(H_t)\) be the singular values of the estimated 
homography \(H_t\). The frame-wise distortion (higher is better) is
\begin{equation}
  D_t = \frac{\sigma_{\min}(H_t)}{\sigma_{\max}(H_t)} \in (0,1].
\end{equation}
The global distortion metric \(D\) is defined as the minimum value across all frames, ensuring robustness against 
outliers:
\begin{equation}
  \label{eq:distortion}
  D = \min_t D_t.
\end{equation}

\paragraph{Stability Score (S)}  
The stability score is defined by frequency-domain analysis of the estimated camera trajectory. The translation and 
rotation components are extracted as 1D temporal signals. The stability score is defined as the proportion of 
low-frequency energy (2nd–6th order frequencies, excluding the DC component). Formally, let \(X(f)\) denote the 
Fourier spectrum of the trajectory signal and \(B_{\text{low}}\) the low-frequency band:
\begin{equation}
  \label{eq:stability}
  S = \frac{\sum_{f \in B_{\text{low}}} |X(f)|^2}{\sum_{f} |X(f)|^2}.
\end{equation}

\noindent \textbf{Implementation details.} Trajectories are accumulated frame by frame using relative homographies, 
from which translation and rotation (approximated by \(\arctan(s_x/s_y)\)) are extracted. FFT is applied to both 
translation and rotation sequences. After removing the DC component, the energy ratio of the first five non-zero 
frequencies (2nd–6th) is computed. The final stability score is the minimum of the translation-based and 
rotation-based ratios, yielding a strict evaluation.

\paragraph{PSNR}  
Peak Signal-to-Noise Ratio (PSNR) is used to measure reconstruction quality in videos and images:
\begin{equation}
  \label{eq:psnr}
  \mathrm{PSNR} = 10 \cdot \log_{10} \Big( \frac{ \mathrm{MAX}_I^2 }{ \mathrm{MSE} } \Big),
\end{equation}
where \(\mathrm{MAX}_I\) is the maximum possible pixel value (255 for 8-bit images). The mean squared error 
(MSE) between a reference frame \(I_i\) and a test frame \(K_i\) with \(N\) total pixels is
\begin{equation}
  \label{eq:mse}
  \mathrm{MSE} = \frac{1}{N} \sum_{i=1}^{N} (I_i - K_i)^2.
\end{equation}


\begin{figure*}[t]
  \centering
  \includegraphics[width=0.9\textwidth]{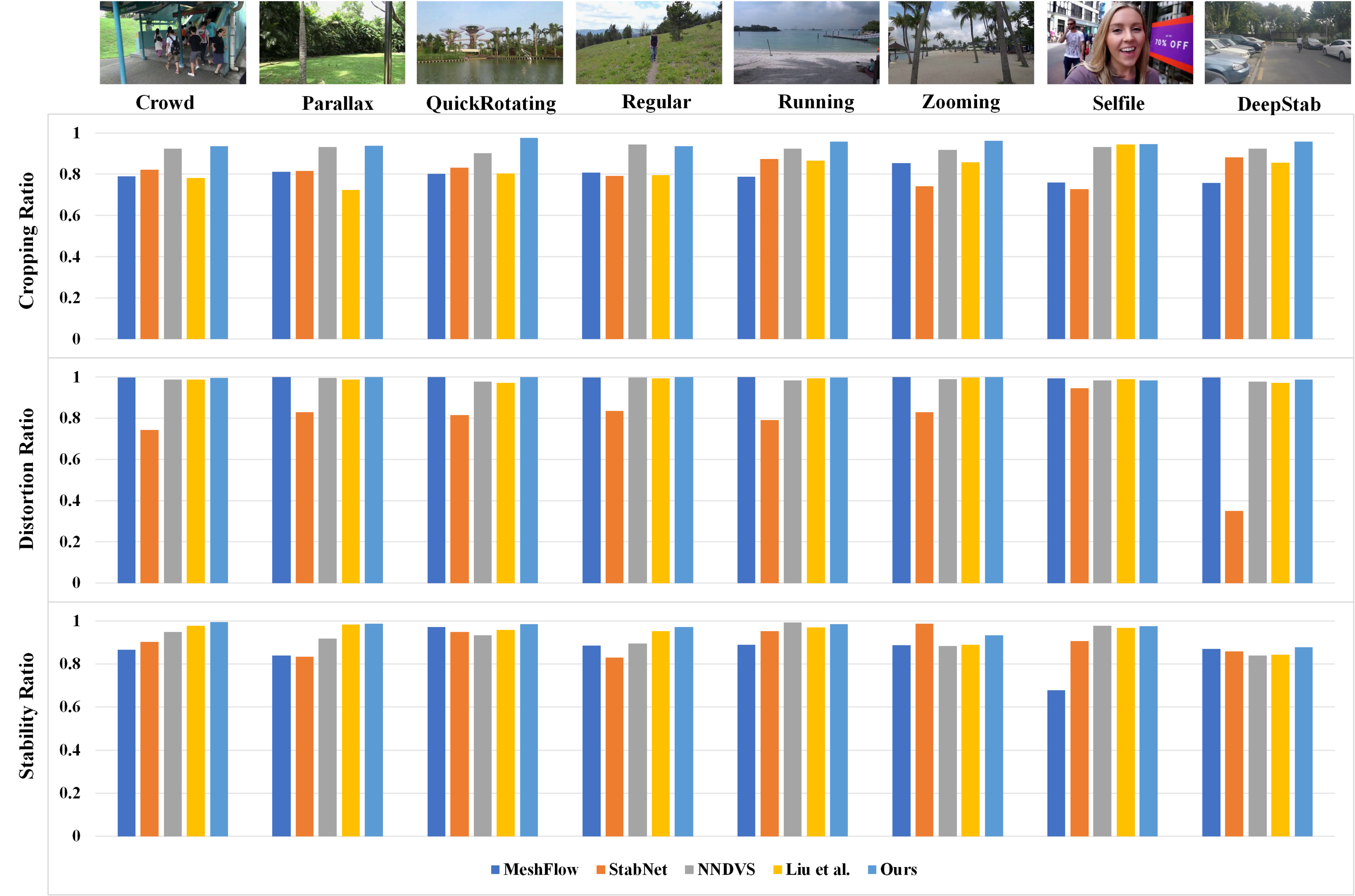}
  \caption{Per-scene quantitative evaluation.}
  \label{fig:perScene}
\end{figure*}

\section{Quantitative Evaluation}
Fig.~\ref{fig:perScene} presents the per-scene evaluation results for our method and other online approaches on 
the NUS~\cite{r20}, Selfie~\cite{r11}, and DeepStab~\cite{r10} datasets. Both our method and representative 2D-based 
full-frame methods~\cite{r5,r21,r24} achieve the maximum cropping ratio of 1, indicating that the stabilized videos 
retain the full field of view without cropping. In terms of distortion and stability, our method achieves performance 
comparable to the state-of-the-art 3D-based method~\cite{r28}, while outperforming 2D-based methods~\cite{r5,r21,r24,r23}. 
In summary, the proposed method demonstrates both effectiveness and robustness across diverse scenarios.

\section{User Study}
\begin{figure}[t]
  \centering
  \includegraphics[width=0.48\textwidth]{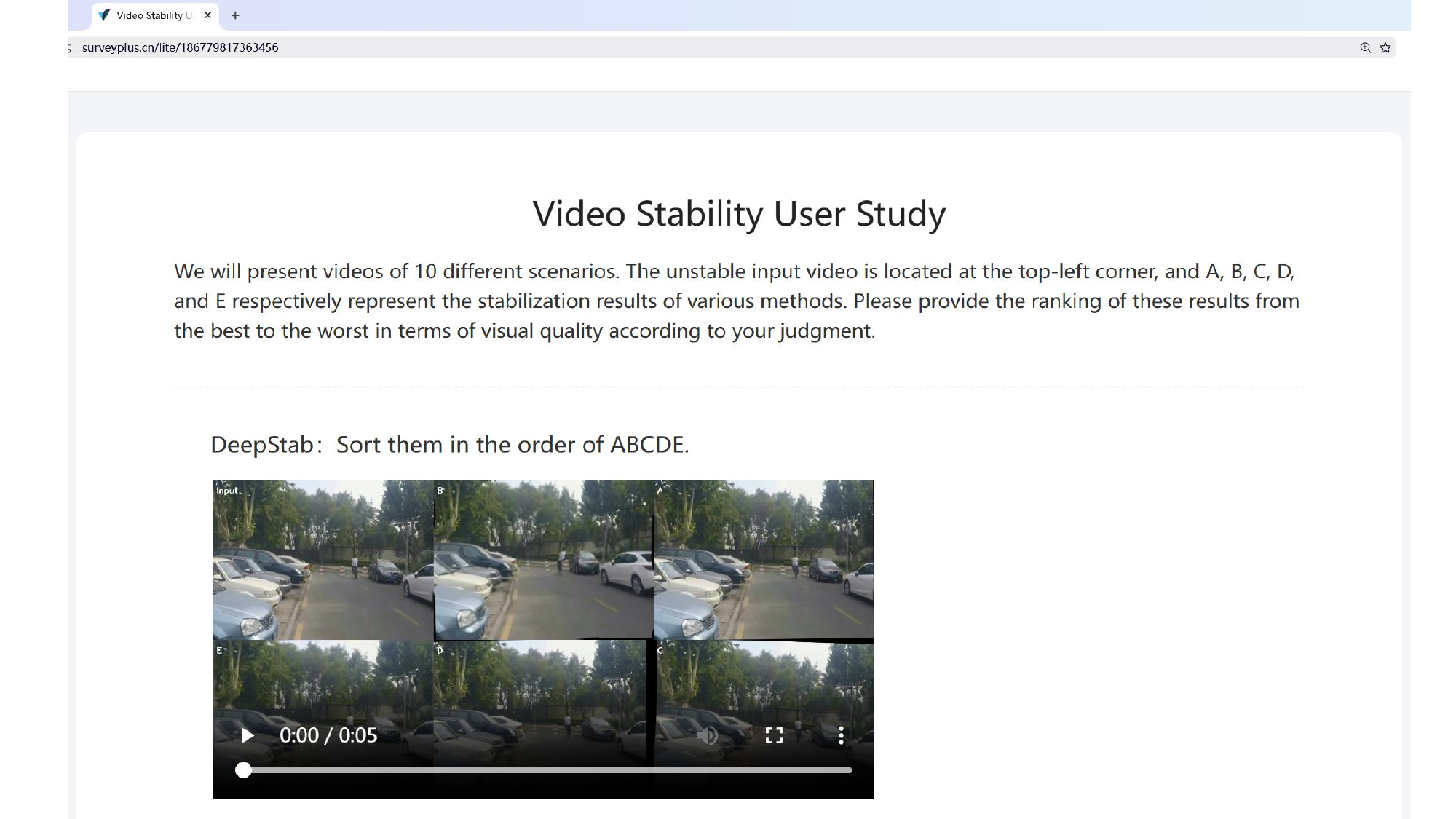}
  \caption{Example question from the user study survey.}
  \label{fig:example_question}
\end{figure}

We conducted a user study via an online SurveyPlus questionnaire. From the five datasets described in the paper, 
we randomly selected two video sequences from each, resulting in ten short video clips in total. The original 
unstable videos were synchronized and displayed side-by-side with stabilized outputs generated by four 
state-of-the-art online methods and our approach. The stabilized videos were arranged in randomized order, with 
each video potentially appearing in either central or peripheral positions. Participants could replay the videos 
to ensure a fair comparison. Finally, they were asked to rank the visual quality from best to worst, with the 
best video assigned 5 points, followed by 4, 3, 2, and 1 point for the worst. 
Figure~\ref{fig:example_question} illustrates an example of our questionnaire design.

\section{Additional Visualization Results}

\paragraph{Motion Visualization}  
Fig.~\ref{fig:visualize_motionpro} illustrates the sparse optical-flow motion guided by the initial keypoints and 
the grid-wise motion obtained after propagation with our \textbf{EfficientMotionPro} module. After propagation, the 
estimated motion remains consistently structured and is less affected by dynamic objects and noise.

\paragraph{Trajectory Visualization}  
\begin{figure}[t]
  \centering
  \includegraphics[width=0.45\textwidth]{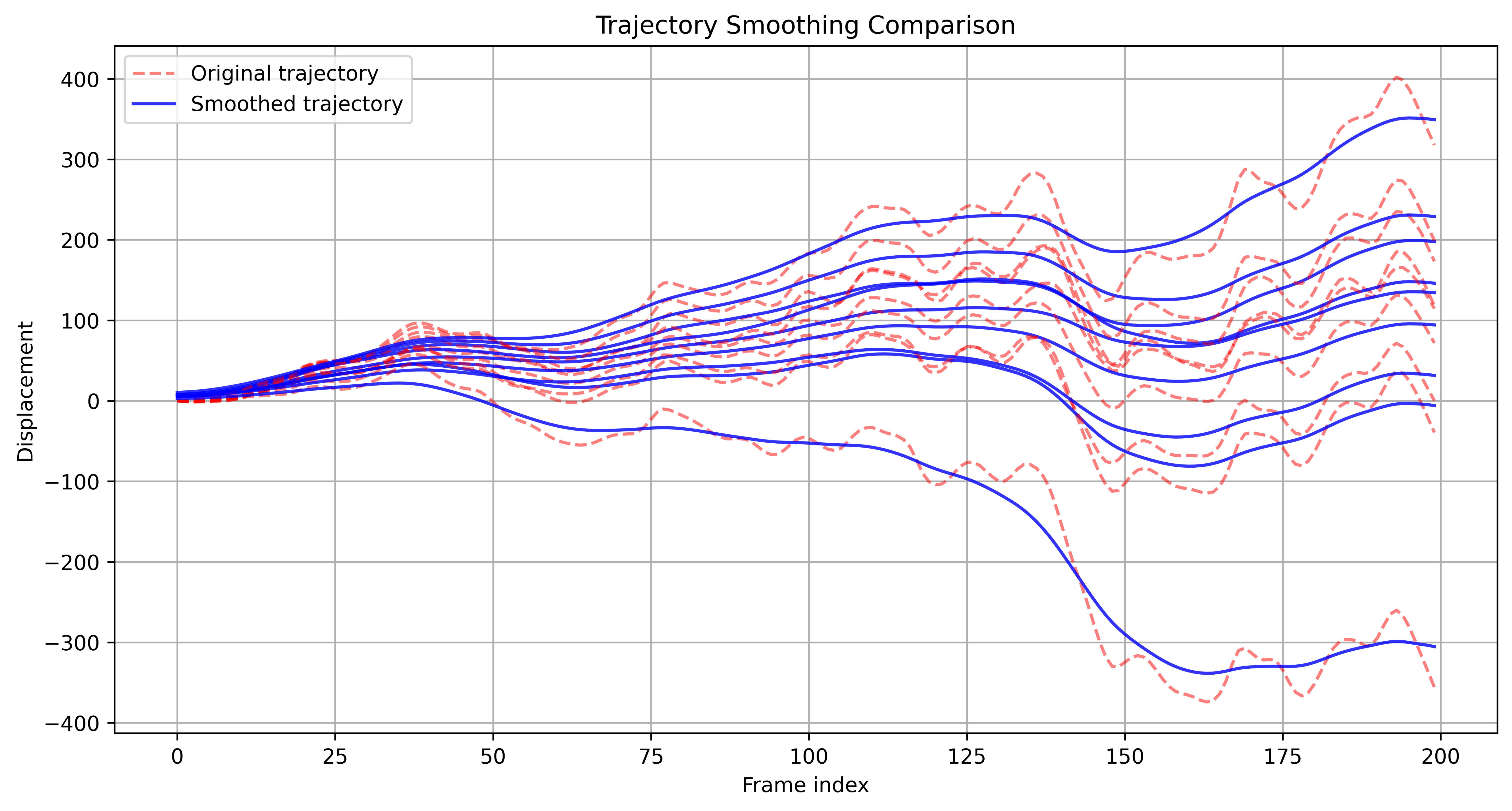}
  \caption{Visualization of mesh vertex trajectories during online smoothing. Six vertices are randomly selected, 
  showing comparisons between unsmoothed trajectories after motion propagation and those refined by our 
  \textbf{OnlineSmoother}.}
  \label{fig:visualize_trajectories}
\end{figure}

\begin{figure}[t]
  \centering
  \includegraphics[width=0.45\textwidth]{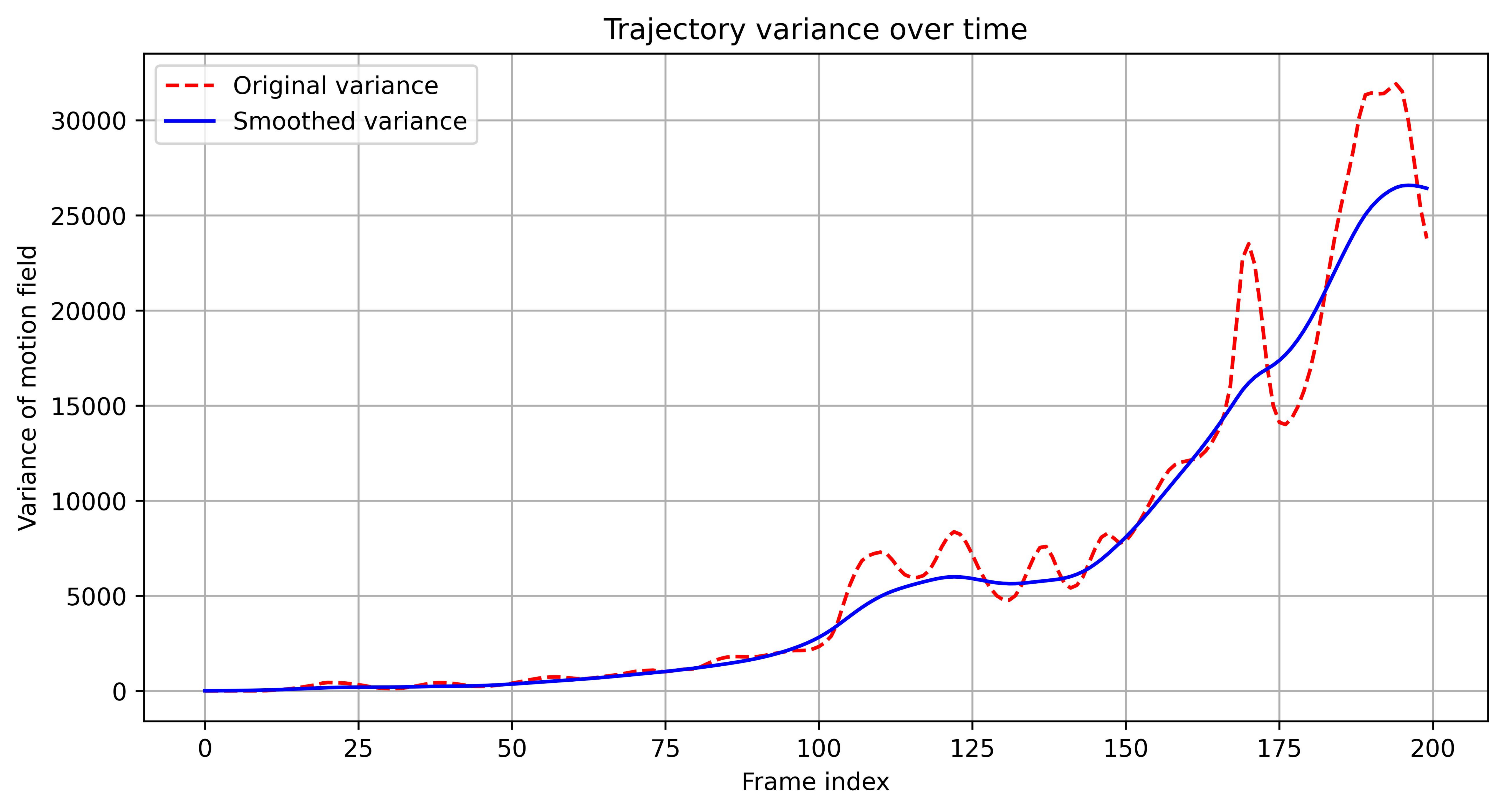}
  \caption{Comparison of trajectory variance before and after applying \textbf{OnlineSmoother}, demonstrating its 
  effectiveness in suppressing fluctuations.}
  \label{fig:variance}
\end{figure}

\begin{figure}[t]
  \centering
  \includegraphics[width=0.45\textwidth]{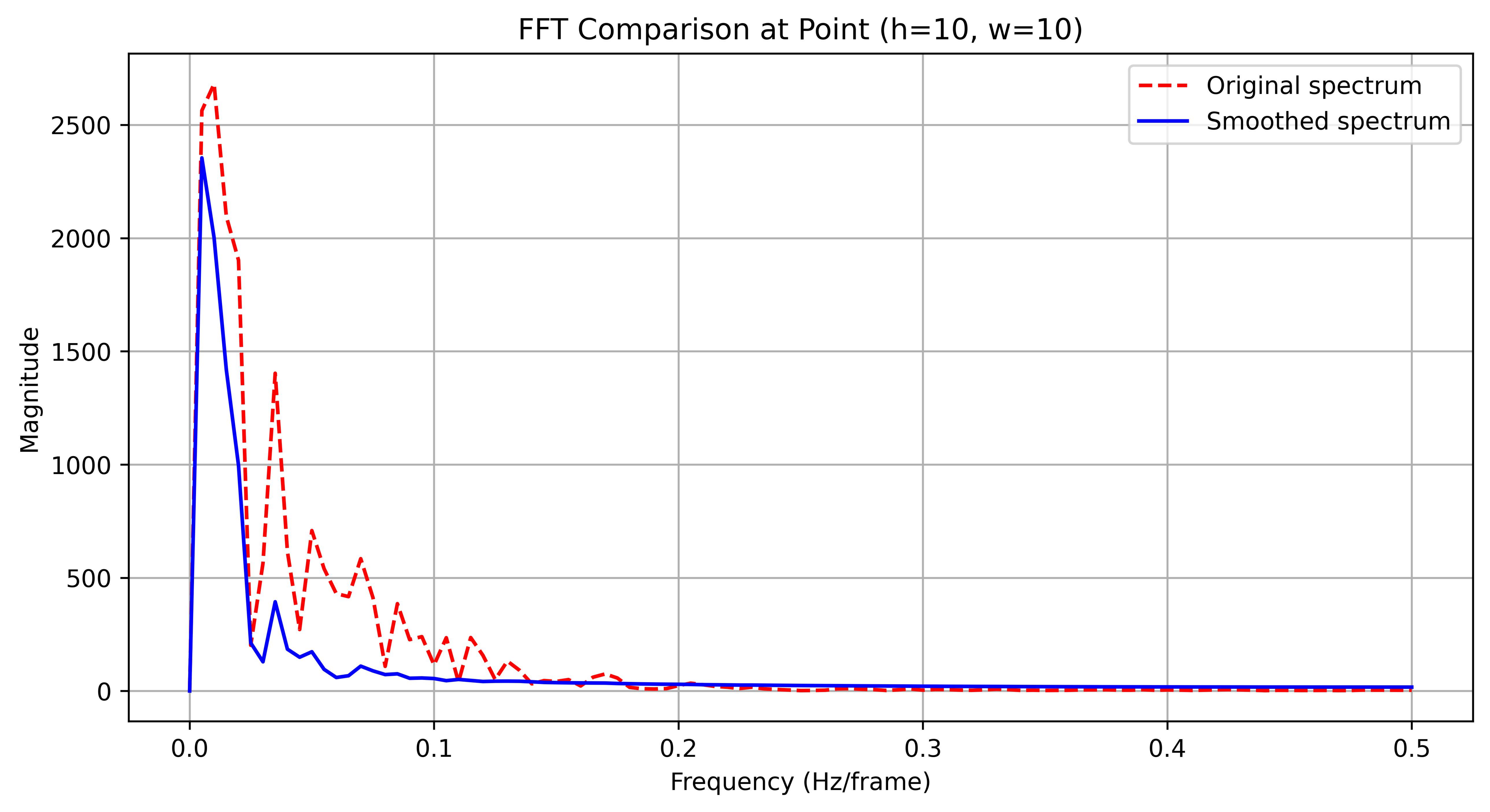}
  \caption{Frequency-domain analysis of trajectories via FFT, highlighting that \textbf{OnlineSmoother} effectively 
  attenuates high-frequency noise.}
  \label{fig:fft_comparison}
\end{figure}

As illustrated in Figs.~\ref{fig:visualize_trajectories}, \ref{fig:variance}, and \ref{fig:fft_comparison}, we 
visualize mesh-vertex trajectories during online smoothing, comparing trajectories obtained after motion propagation 
without \textbf{OnlineSmoother} to those refined by \textbf{OnlineSmoother}. For clarity, we randomly select six 
mesh vertices to display their smoothed trajectories. We also plot trajectory variance and FFT spectra to further 
highlight the effectiveness of our OnlineSmoother module.

\paragraph{Keypoint Density Visualization}  

\begin{figure}[t]
\centering
\includegraphics[width=0.9\linewidth]{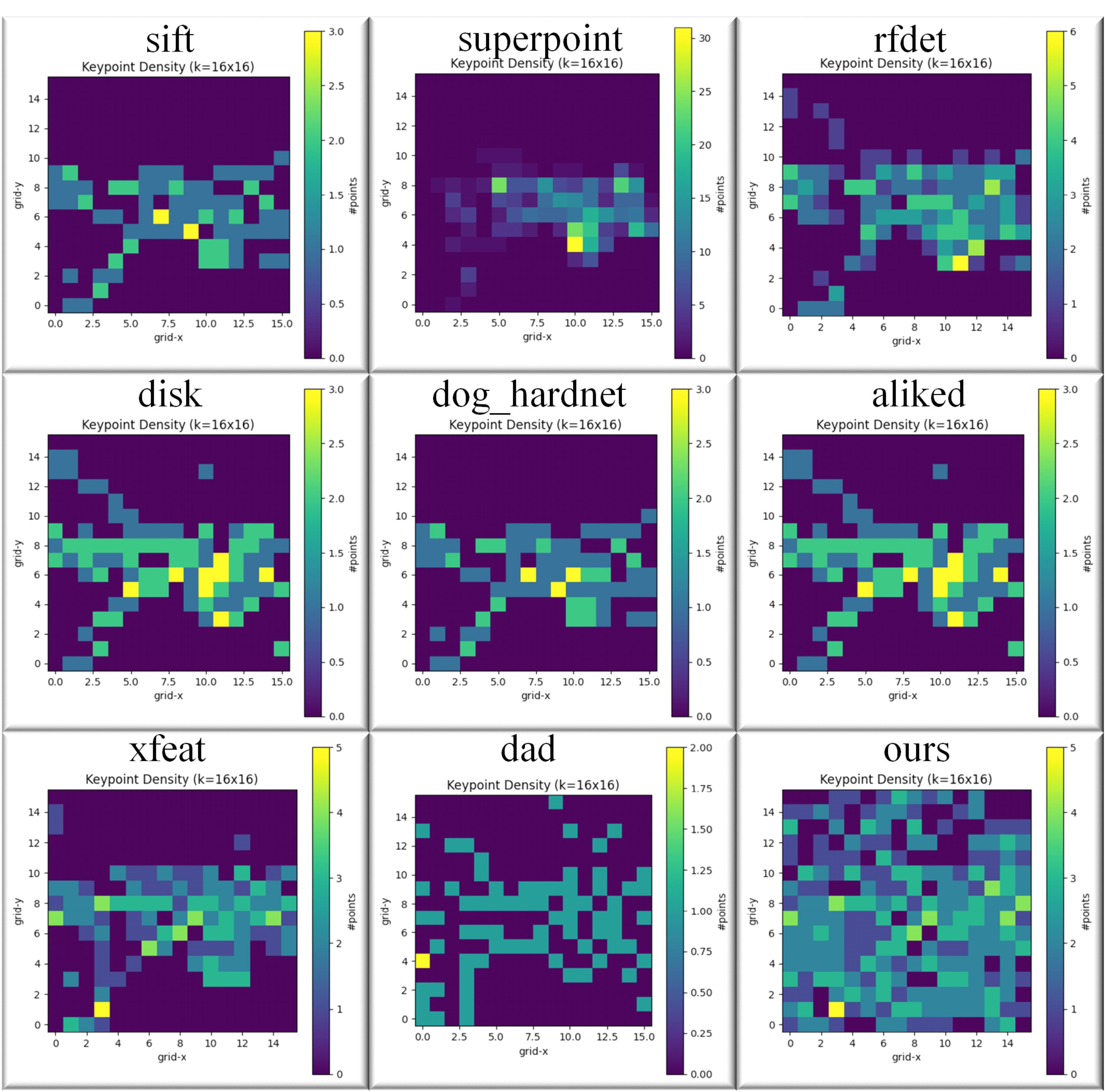}
\caption{Keypoint density heatmaps of representative detectors (SIFT~\cite{r18}, SuperPoint~\cite{r29}, 
RFdet~\cite{r30}, DISK~\cite{r31}, DoG-HardNet~\cite{r62}, ALIKED~\cite{r33}, XFeat~\cite{r34}, DAD~\cite{r35}, and 
Ours) on a sample scene. Our collaborative detector achieves more uniform coverage across the image grid.}
\label{fig:keypoint_density}
\end{figure}

To further illustrate the advantage of our collaborative detector, we visualize keypoint density maps of 
different detectors in Fig.~\ref{fig:keypoint_density}. The heatmaps are computed over a $16 \times 16$ grid. 
As shown, traditional and learning-based detectors often produce highly clustered distributions 
(e.g., SIFT~\cite{r18} and SuperPoint~\cite{r29}), whereas our collaborative method yields more uniform spatial 
coverage, providing a stronger foundation for downstream geometric estimation.

\begin{figure*}[t]
\centering
\includegraphics[width=1\linewidth]{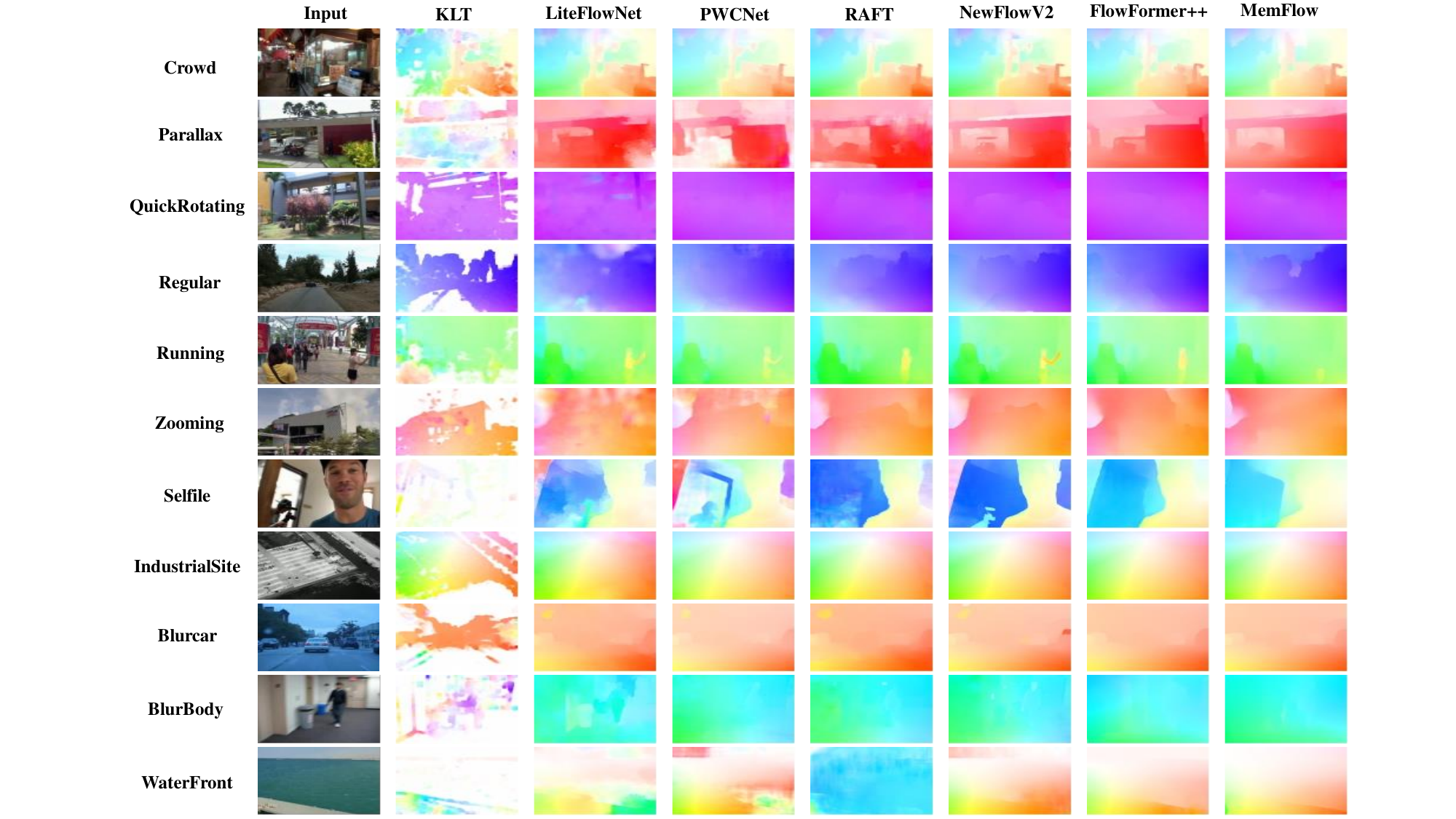}
\caption{Comparison of representative optical-flow algorithms on diverse examples. We report results from several 
state-of-the-art methods~\cite{r66,r42,r41,r43,r47,r46,r48}, among which MemFlow~\cite{r48} is adopted in our 
experiments for its balance of accuracy and memory efficiency.}
\label{fig:OpticalFlow}
\end{figure*}

\paragraph{Optical Flow Visualization}  
Fig.~\ref{fig:OpticalFlow} presents the results of several representative and state-of-the-art optical-flow 
algorithms, including~\cite{r66,r42,r41,r43,r47,r46,r48}. In our experiments, we adopt MemFlow~\cite{r48}, 
which achieves high accuracy while being memory-efficient.

\section{Video Results}
In the supplementary material, we provide a 3-minute comparison video of our online stabilization method against other online methods. This video offers an intuitive view of our method's behavior across diverse scenarios.

\section{Limitations}
Our method achieves unsupervised online video stabilization through explicit motion estimation and trajectory smoothing. 
However, there are two main limitations. First, it relies on existing optical-flow estimators, which may be less accurate in complex scenes. Second, to improve visual quality, we currently rely on a post-processing frame outpainting step to handle black borders; due to its computational cost, it is not integrated into the online pipeline. 
To address these issues, we plan to (i) explore more accurate yet efficient optical-flow models, and (ii) develop lighter, online-friendly outpainting techniques that improve edge quality without sacrificing real-time performance.

{
    \small
    \bibliographystyle{ieeenat_fullname}
    \bibliography{main}
}
\end{document}